\documentclass[]{heygen_research}

\usepackage[toc,page,header]{appendix}
\usepackage{natbib}

\usepackage{amsmath}
\usepackage{amssymb}
\usepackage{dsfont}

\usepackage{cleveref}

\usepackage{subcaption}

\usepackage{xargs}
\usepackage{todonotes}

\usepackage{multirow}

\usepackage{algorithm}
\usepackage{algorithmic}

\usepackage{array}
\usepackage{tabularx}

\usepackage{tcolorbox}

\usepackage{adjustbox}

\usepackage{bm}
\usepackage{mathrsfs}

\usepackage{adjustbox}
\usepackage{multicol}
\usepackage{changepage}
\usepackage{graphicx}
\usepackage{array}
\usepackage{hyperref}

\usepackage{minitoc}
\usepackage{amsmath,amsfonts,bm}

\def\eqref#1{equation~\ref{#1}}

\def\1{\bm{1}}

\DeclareMathAlphabet{\mathsfit}{\encodingdefault}{\sfdefault}{m}{sl}
\SetMathAlphabet{\mathsfit}{bold}{\encodingdefault}{\sfdefault}{bx}{n}

\newcommand{\ccmethod}{TransVLM}

\title{\ccmethod{}: A Vision-Language Framework and Benchmark for Detecting Any Shot Transitions}
\headertitle{\ccmethod{}: VLM for Shot Transition Detection}

\author[1,2,*]{Ce Chen}
\author[2]{Yi Ren}
\author[2]{Yuanming Li}
\author[2]{Viktor Goriachko}
\author[2]{Zhenhui Ye}
\author[2,3,*]{Zujin Guo}
\author[2]{Zhibin Hong}
\author[1,\dagger]{Mingming Gong}

\affiliation[1]{University of Melbourne}
\affiliation[2]{HeyGen Research}
\affiliation[3]{Nanyang Technological University}

\abstract{
Traditional Shot Boundary Detection (SBD) inherently struggles with complex transitions by formulating the task around isolated cut points, frequently yielding corrupted video shots. We address this fundamental limitation by formalizing the Shot Transition Detection (STD) task. Rather than searching for ambiguous points, STD explicitly detects the continuous temporal segments of transitions. To tackle this, we propose \ccmethod{}, a Vision-Language Model (VLM) framework for STD. Unlike regular VLMs that predominantly rely on spatial semantics and struggle with fine-grained inter-shot dynamics, our method explicitly injects optical flow as a critical motion prior at the input stage. Through a simple yet effective feature-fusion strategy, \ccmethod{} directly processes concatenated color and motion representations, significantly enhancing its temporal awareness without incurring any additional visual token overhead on the language backbone. To overcome the severe class imbalance in public data, we design a scalable data engine to synthesize diverse transition videos for robust training, alongside a comprehensive benchmark for STD. Extensive experiments demonstrate that \ccmethod{} achieves superior overall performance, outperforming traditional heuristic methods, specialized spatiotemporal networks, and top-tier VLMs.
}
\checkdata[Project Page]{\url{https://chence17.github.io/TransVLM/}}

\begin{document}

\maketitle
\let\thefootnote\relax\footnotetext{$*$\,Work done during an internship at HeyGen Research. $\dagger$\,Corresponding author.}
\let\thefootnote\relax\footnotetext{For more related research, please visit \href{https://www.heygen.com/research}{HeyGen Research} and \href{https://www.heygen.com/research/avatar-v-model}{HeyGen Avatar-V}.}

\section{Introduction}
\label{sec:intro}

\begin{figure}[t]
	\centering
	\includegraphics[width=\linewidth]{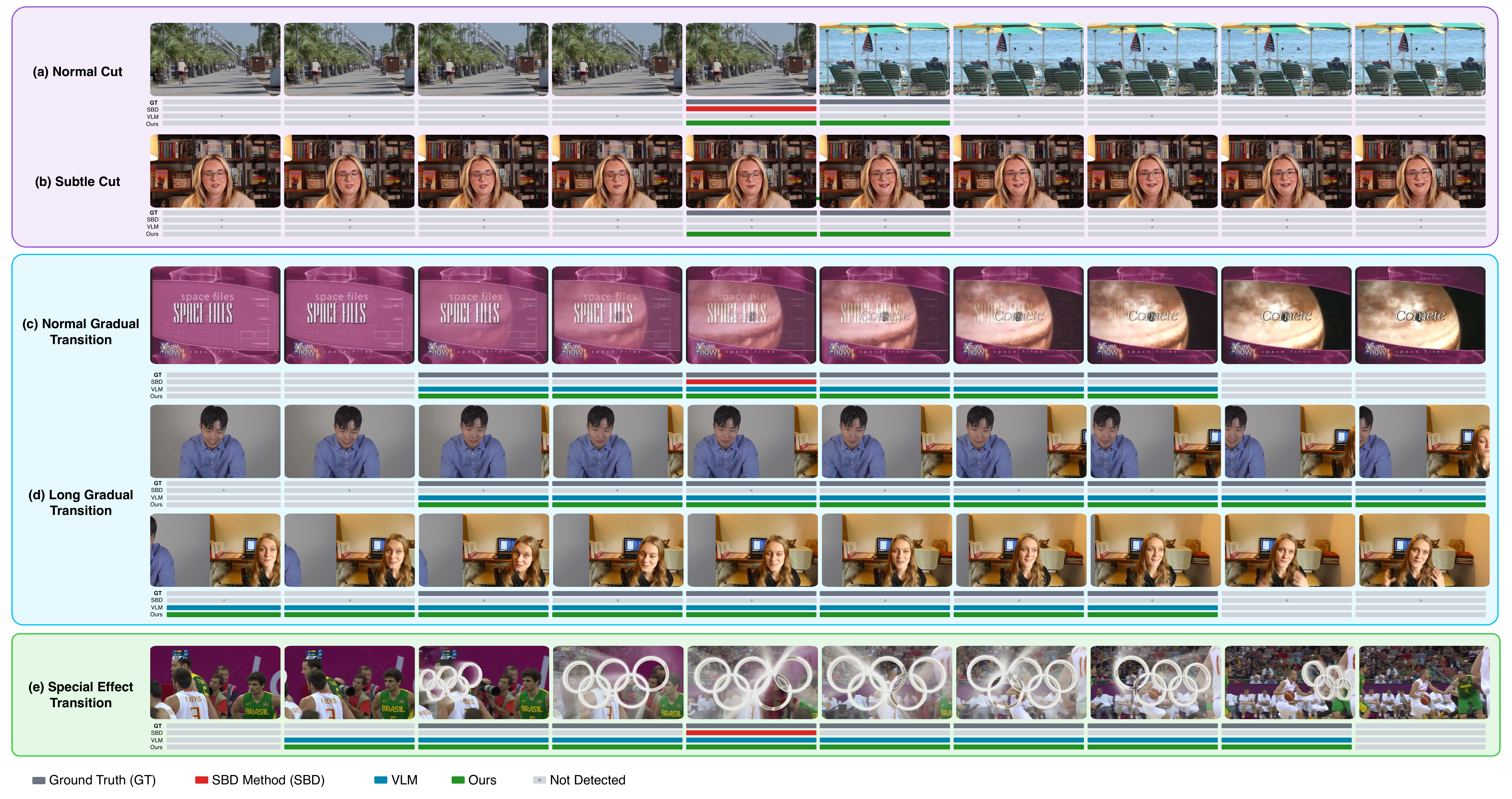}
    \caption{
        \textbf{Limitations of existing shot transition detection methods.} Predicted transitions are denoted by colored lines: \textbf{\textcolor{gray}{gray}} for ground truth, \textbf{\textcolor{red}{red}} for a state-of-the-art SBD method (AutoShot \cite{zhu2023autoshot}), \textbf{\textcolor{blue}{blue}} for a general-purpose top-tier VLM (Qwen3-VL \cite{bai2025qwen3}), and \textbf{\textcolor{green}{green}} for our proposed \ccmethod{}. While SBD models excel at detecting normal cuts (a) but fail on gradual and special transitions (c, d, e), VLMs can perceive gradual and special transitions but miss normal cuts. Neither of them can detect the subtle cuts (b). In contrast, \ccmethod{} robustly detects all these transitions.
    }
    \label{fig:teaser}
\end{figure}

Precisely detecting the start and end timestamps of shot transitions is a critical prerequisite for modern video processing. By identifying these accurate timestamps, models are shielded from the noise of inter-shot transitions, making this task a foundational pillar for a wide range of downstream applications. In video retrieval, processing clean shots strictly prevents the semantic blending of distinct scenes, thereby ensuring robust cross-modal matching performance \cite{bain2021frozen, luo2022clip4clip}. In video understanding tasks \cite{wang2022internvideo, tong2022videomae, kwon2023efficient, xu2021videoclip, actionatlas2024}, such as action recognition and dense captioning, accurate shot transition timestamps guarantee that models learn coherent spatiotemporal dynamics without being disrupted by abrupt visual shifts. Most importantly, in the recent surge of text-to-video (T2V) generation \cite{blattmann2023stable, brooks2024videogeneration, ijcnlp2025transition, soradetector2024, vista2025veo3, zhang2025waver}, detecting shot transitions is strictly necessary for data curation and label preparation. If the transition information in the training labels is insufficient or inaccurate, it actively causes generative models to produce unintended shot transitions in the generated videos.

To achieve this, two primary paradigms currently exist: traditional Shot Boundary Detection (SBD) methods and Vision-Language Models (VLMs). However, in practice, both exhibit critical flaws. We observe that while existing SBD methods perform adequately on normal abrupt cuts (Figure \ref{fig:teaser}(a)), their performance degrades significantly when encountering complex transitions. For instance, when processing gradual transitions (e.g., dissolves, fades, and wipes), traditional SBD methods often fail to determine precise boundaries, resulting in extracted shots contaminated by dirty transitional frames (Figure \ref{fig:teaser}(c)). Furthermore, these methods frequently miss subtle cuts (cuts with minor content changes), long gradual transitions, and special effect transitions (Figure \ref{fig:teaser}(b, d, e)). Consequently, these corrupted shots severely degrade downstream processing. Conversely, our preliminary experiments reveal that while VLMs exhibit promising performance on complex gradual and special effect transitions (Figure \ref{fig:teaser}(c, d, e)), they struggle significantly with abrupt cuts (Figure \ref{fig:teaser}(a, b)). 

To fundamentally resolve these limitations, we reformulate the traditional SBD paradigm and propose the Shot Transition Detection (STD) task. Instead of focusing solely on content variations between adjacent frames like conventional SBD, STD emphasizes the intrinsic spatiotemporal patterns of the transitions connecting different shots. This includes capturing both content and motion dynamics across shots, rather than merely focusing on frame-to-frame content differences. Our detection targets logically shift from isolated cut points to continuous shot transition segments. To standardize evaluation in this newly formalized task, we construct a large-scale, high-quality STD benchmark with multi-dimensional metrics.

\begin{figure}[t]
	\centering
	\includegraphics[width=\textwidth]{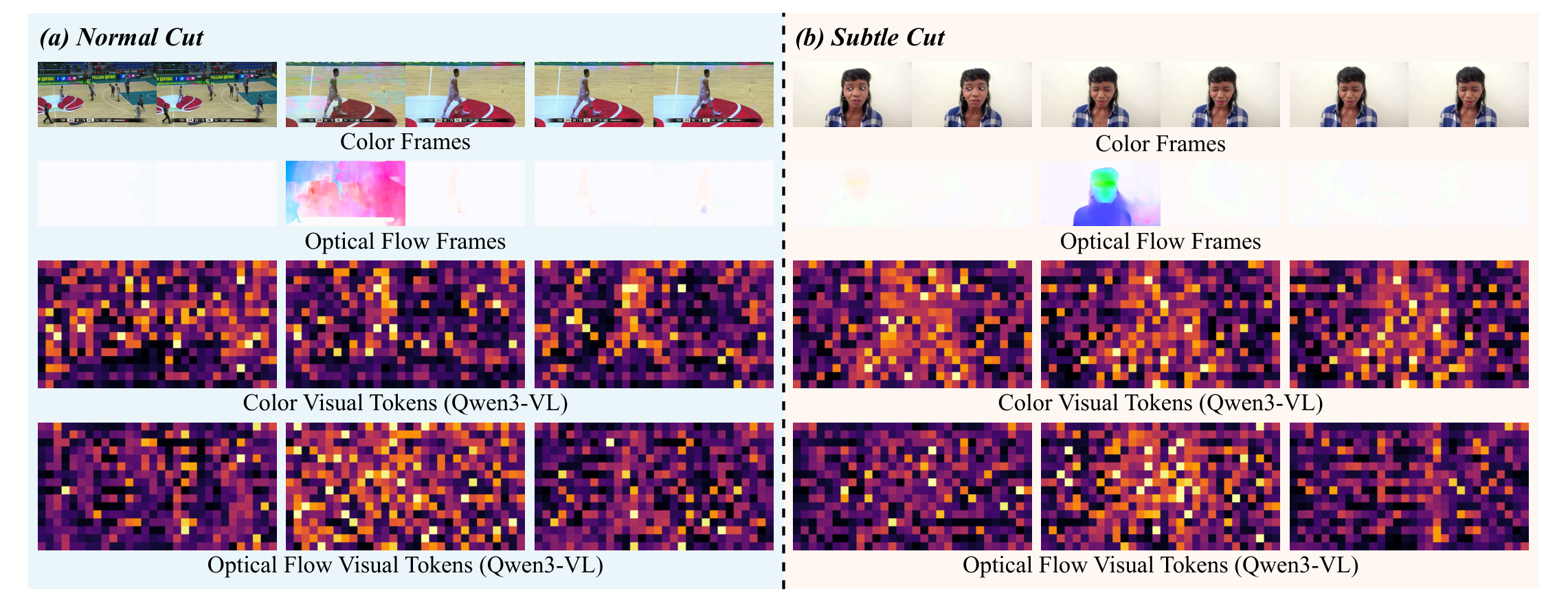}
    \caption{
        \textbf{Visual token visualization for color vs. optical flow.} By capturing inter-frame changes, optical flow serves as a robust motion prior. For both (a) normal and (b) subtle cuts, the visual tokens derived from optical flow exhibit significantly sharper response contrast at the transition frames compared to standard color tokens. Visualizations are extracted from Qwen3-VL \cite{bai2025qwen3} (temporal downsampling stride of 2).
        }
    \label{fig:optical_flow}
\end{figure}

To address the respective defects of SBD methods and VLMs on the STD task, we propose the \ccmethod{} framework. We attribute the suboptimal VLM performance on normal cuts to the sparse, low-frame-rate inputs (e.g., 5 FPS) conventionally required to prevent out-of-memory errors and severe inference latency. Such aggressive temporal downsampling makes detecting instantaneous cuts highly challenging. To overcome this, we introduce a temporal sliding-window strategy. By partitioning the video stream into fixed-length, overlapping segments for sequential inference and subsequently merging the local outputs into continuous global predictions, we enable the VLM to reliably focus on and detect fast-changing cuts without exceeding memory constraints.

Furthermore, for subtle cuts, the failure stems from the VLM's inherent bias toward static spatial semantics rather than fine-grained inter-shot motion dynamics. We discover that explicitly providing optical flow is exceptionally effective at capturing these missing motion priors. As illustrated in Figure \ref{fig:optical_flow}, for both normal and subtle cuts, the temporal variations at the exact transition frames---evident in both the visual frames and the final visual tokens processed by the language model---are significantly more distinguishable in the optical flow modality than in standard color frames. To seamlessly inject this optical flow into the network without inflating the computational burden, we design a simple yet effective feature-fusion strategy, which integrates motion representations while strictly maintaining the same sequence length of visual tokens. Finally, to mitigate the critical scarcity of annotated transitions, our specially designed data engine is leveraged to automatically synthesize massive, high-quality training data.

Our main contributions are summarized as follows:
\begin{itemize}
    \item We reformulate the traditional SBD paradigm and propose the novel Shot Transition Detection (STD) task, alongside a comprehensive benchmark equipped with multi-dimensional metrics specifically tailored for STD.
    \item We propose \ccmethod{}, an efficient vision-language framework that explicitly integrates optical flow as a motion prior. Our feature-fusion strategy significantly enhances the capability of detecting abrupt cuts without incurring any additional visual token overhead.
    \item We design a scalable data engine capable of automatically synthesizing diverse transition videos, providing high-quality supervision and enabling robust VLM training for the STD task.
    \item Extensive experiments demonstrate that \ccmethod{} achieves superior overall performance on the proposed STD benchmark, substantially outperforming traditional heuristic methods, specialized spatiotemporal networks, and top-tier general-purpose VLMs.
\end{itemize}
\section{Related Work}
\label{sec:relwo}

\noindent\textbf{Shot Boundary Detection.} Traditional SBD approaches \cite{abdulhussain2018methods, kar2024video} rely on heuristic algorithms (e.g., PySceneDetect \cite{pyscenedetect}, ECR \cite{zabih1995feature}) that compute low-level visual differences. While computationally inexpensive, they are highly sensitive to local illumination changes and fast motion, leading to frequent false negatives. Recent deep learning architectures \cite{xu2016shot, wu2019two, tong2015cnn}, such as the TransNet series \cite{soucek2019transnet, soucek2020transnet} and AutoShot \cite{zhu2023autoshot}, utilize Convolutional Neural Networks (CNNs) to significantly improve detection performance. However, by formulating the task as a frame-wise binary classification, these methods exclusively target isolated cut points. Consequently, their discrete probability outputs falter on complex gradual transitions, frequently yielding corrupted shots containing dirty frames. 

\noindent\textbf{Video Understanding.} Recent advancements in video understanding have evolved from specialized spatiotemporal architectures (e.g., I3D \cite{carreira2017quo}, SlowFast \cite{feichtenhofer2019slowfast}, and VideoMAE \cite{tong2022videomae}) to powerful general-purpose Vision-Language Models (VLMs), such as Video-LLaVA \cite{lin2023video}, the Qwen-VL series \cite{wang2023qwen, qwen2vl}, and Gemini \cite{team2023gemini}. While their robust cognitive capabilities are theoretically well-suited for modeling content and motion variations across shots, directly applying off-the-shelf VLMs to the STD task yields imprecise transition segments. This failure stems from two core limitations. First, VLMs inherently prioritize static spatial semantics over fine-grained inter-shot dynamics. Second, their reliance on sparse, low-frame-rate inputs to prevent memory overload critically restricts their ability to detect instantaneous cuts. 


\section{Task Formulation}
\label{sec:task_formulation}

Given a video $V$ consisting of $N$ frames, traditional SBD approaches formulate the task as a frame-wise binary classification problem. The output is typically represented as a sequence of frame-level probabilities, $\mathcal{P}_{frame} = \{p_1, p_2, \dots, p_N\}$, where each $p_i$ indicates the likelihood of the $i$-th frame being a shot boundary. Boundaries are subsequently extracted by applying a predefined heuristic threshold. However, these SBD models predominantly focus on detecting isolated cut points while entirely neglecting the intrinsic spatiotemporal patterns of the transitions. Consequently, when processing complex transitions (e.g., dissolves and wipes), the predicted probabilities for transitional frames often lack salient contrast compared to intra-shot frames. This ambiguity makes it nearly impossible to establish a robust threshold capable of cleanly separating transition segments from pure video shots.

In contrast, we formally define our proposed STD as a transition detection task. Rather than assigning independent probability scores to individual frames, STD requires the model to explicitly detect the complete transition segments, thereby holistically capturing both content and motion dynamics across shots. Each transition segment is formally defined as a temporal tuple comprising its precise start and end times. Thus, the prediction is formulated as a set of discrete temporal segments, $\mathcal{P}_{segment} = \{(s_1, e_1), (s_2, e_2), \dots, (s_M, e_M)\}$, where $s_i$ and $e_i$ denote the start and end timestamps of the $i$-th transition, respectively. This representation naturally unifies the definition of abrupt cuts (where $s_i \approx e_i$) and gradual or special effect transitions (where $s_i < e_i$). 

Crucially, this segment-level formulation offers two distinct advantages. First, the tuple-based format can be effortlessly serialized into natural language tokens, seamlessly aligning the STD task with VLMs. Second, by explicitly forcing the model to predict the continuous temporal span of a transition between shots rather than isolated cut points of a shot, this formulation serves as a strong inductive bias. It actively encourages the model to capture the intrinsic spatiotemporal patterns of the transition dynamics themselves, rather than over-fitting to ambiguous boundary frames.

\section{\ccmethod{} Framework}
\label{sec:method}

\begin{figure*}[t]
    \centering
    \includegraphics[width=\textwidth]{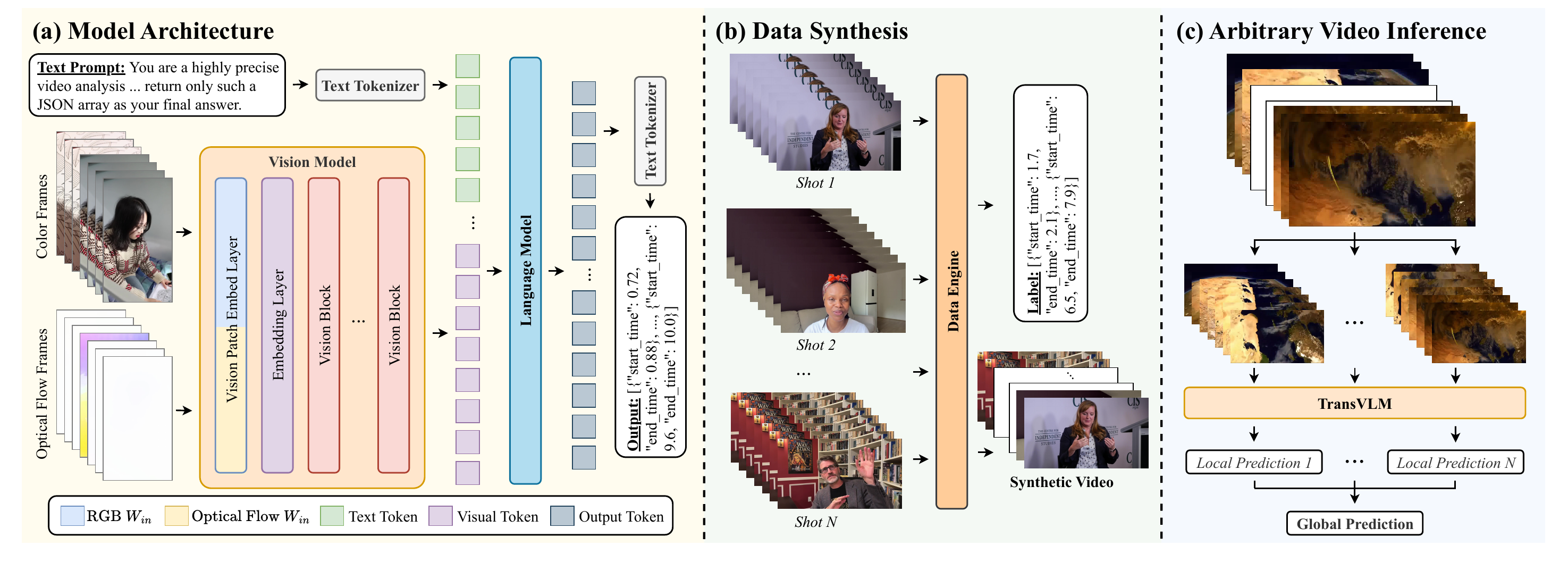}
    \caption{
    	\textbf{\ccmethod{} Framework.} Our proposed framework comprises three core components. \textbf{(a) Model Architecture:} We explicitly inject optical flow as a motion prior via a parameter-efficient strategy. By exclusively expanding the input projection weights ($W_{in}$) of the Vision Patch Embed Layer, the model directly processes concatenated frames of color and optical flow. Crucially, this extracts joint appearance-motion representations without inflating the visual token sequence length, thereby incurring zero additional computational burden on the language model. \textbf{(b) Data Synthesis:} Given an arbitrary sequence of clean shots, our scalable data engine automatically synthesizes videos with diverse transitions, simultaneously generating precisely aligned segment-level JSON labels for model training. \textbf{(c) Arbitrary Video Inference:} To process videos of unconstrained duration without causing memory overflow, we employ a temporal sliding-window strategy. The input stream is partitioned into overlapping segments to generate local predictions, which are subsequently aggregated into a continuous global prediction via temporal Non-Maximum Suppression (NMS).
    }
    \label{fig:architecture}
\end{figure*}

In this section, we detail the proposed \ccmethod{} framework for STD task. We first introduce the overall model architecture, specifically focusing on the feature-fusion mechanism for integrating optical flow and the zero-padding weight initialization strategy (Section \ref{method:model_arch}). Subsequently, we elaborate on the data engine, the quality-aware mixed sampling strategy, and the arbitrary video inference pipeline (Section \ref{method:model_train}).

\subsection{Model Architecture}
\label{method:model_arch}

Existing Vision-Language Models (VLMs) generally exhibit suboptimal performance on the STD task. To overcome these limitations, we propose \ccmethod{}, a novel multimodal framework tailored for precise temporal transition detection. \ccmethod{} explicitly integrates motion priors via a parameter-efficient feature-fusion mechanism without inflating the visual token sequence length.

While VLMs possess the robust cognitive capacity theoretically required for complex video tasks, their pre-trained vision encoders are inherently biased towards high-level semantics and shot content. However, fine-grained inter-shot motion changes are crucial for the STD task. To leverage this information, we explicitly inject optical flow as a motion prior into the VLM. Since Qwen3-VL \cite{bai2025qwen3} is one of the top-tier open-source VLMs, we employ it as our base backbone for \ccmethod{}. Theoretically, our modifications are model-agnostic and can be applied to any VLM architecture.

As depicted in Figure \ref{fig:architecture}(a), the input to \ccmethod{} consists of both standard color frames and optical flow frames. These two modalities are fused at the input stage and processed by the modified vision model. The resulting visual tokens, alongside the tokenized text prompt instructing the model to output transition segments as a structured JSON array, are fed into the language model to predict the exact start and end timestamps.

\noindent\textbf{Feature-Fusion for Motion Prior.} To inject a strong motion prior into the model, we explicitly extract optical flow between consecutive frames. For an optimal balance between computational efficiency and estimation quality, we utilize NeuFlow v2 \cite{zhang2024neuflowv2,zhang2024neuflow,zhang2025neuflow} as our optical flow extractor. To align the spatial distribution of the motion data with the pre-trained visual distribution expected by the VLM, the raw two-dimensional optical flow fields $(d_x, d_y)$ are mapped into a standardized three-channel color visualization format \cite{baker2011database}, treating the flow magnitude and orientation as color saturation and hue.

Let $\mathbf{I}_{\text{color}} \in \mathbb{R}^{H \times W \times 3}$ denote the original video frame and $\mathbf{I}_{\text{Flow}} \in \mathbb{R}^{H \times W \times 3}$ denote the corresponding visualized optical flow frame, where $H$ and $W$ represent the spatial height and width, respectively. We propose a data-level feature-fusion strategy by concatenating these modalities strictly along the channel dimension:
\begin{equation}
\mathbf{I}_{\text{Fused}} = \text{Concat}(\mathbf{I}_{\text{color}}, \mathbf{I}_{\text{Flow}}) \in \mathbb{R}^{H \times W \times 6}.
\end{equation}

To process this fused tensor, we only modify the first layer of the vision model (i.e., the Vision Patch Embedding layer in Qwen3-VL), symmetrically expanding its input channels from 3 to 6. Crucially, because the output channel dimension of this embedding layer remains unchanged, the resulting sequence length of the visual tokens is strictly identical to that of a standard 3-channel input. This simple yet highly effective design ensures that the integration of the motion prior introduces zero additional computational overhead to the subsequent vision blocks and the language model.

\noindent\textbf{Weight Initialization for Stable Fine-tuning.} Directly expanding the input channels of the pre-trained Vision Patch Embedding layer invalidates its original weight tensor dimensions, shifting from $(C, 3, D_{k}, H_{k}, W_{k})$ to $(C, 6, D_{k}, H_{k}, W_{k})$, where $D_{k}$, $H_{k}$, and $W_{k}$ denote the depth, height, and width of the 3D convolutional kernel, respectively. A naive random initialization or uniform scaling of the newly added optical flow channels risks generating massive noisy gradients during early training epochs, potentially leading to catastrophic forgetting of the highly optimized spatial representations learned during the VLM's pre-training phase.

To guarantee optimization stability, we adopt a strict zero-padding initialization strategy. Let $\mathbf{W}_{\text{in}} \in \mathbb{R}^{C \times 3 \times D_{k} \times H_{k} \times W_{k}}$ denote the pre-trained weights of the original 3-channel convolutional layer. The extended weights $\mathbf{W}^{'}_{\text{in}}$ are initialized as follows:
\begin{equation}
\mathbf{W}^{'}_{\text{in}} = \text{Concat}(\mathbf{W}_{\text{in}}, \mathbf{W}^{0}_{\text{in}}) \in \mathbb{R}^{C \times 6 \times D_{k} \times H_{k} \times W_{k}},
\end{equation}
where $\mathbf{W}^{0}_{\text{in}}$ is a zero-initialized tensor of the exact same shape as $\mathbf{W}_{\text{in}}$. This ensures that at the first training step, the model behaves identically to the pre-trained VLM, allowing it to progressively and safely learn to attend to the newly introduced motion prior without disrupting existing spatial knowledge.

\subsection{Model Training and Inference}
\label{method:model_train}

\begin{table*}[t]
	\centering
	\caption{\textbf{Details of the \ccmethod{} Training Dataset.} Our comprehensive training dataset comprises existing public data (row 1-4),  high-fidelity manual annotated data (row 5), highly scalable synthesized data (row 6) and all training data (row 7).}
	\label{tab:train_data_stats}
	\resizebox{\textwidth}{!}{
		\begin{tabular}{
			>{\raggedright\arraybackslash}p{5.50cm}
			>{\raggedright\arraybackslash}p{2.0cm}
			>{\centering\arraybackslash}p{2.25cm}
			>{\raggedleft\arraybackslash}p{1.25cm}
			>{\raggedleft\arraybackslash}p{1.75cm}
			>{\raggedleft\arraybackslash}p{1.25cm}
			>{\raggedleft\arraybackslash}p{1.75cm}
			>{\raggedleft\arraybackslash}p{1.5cm}
			>{\raggedleft\arraybackslash}p{1.5cm}
			>{\raggedleft\arraybackslash}p{1.25cm}
		}
		\toprule
		\multirow{2}{*}{\textbf{Dataset}} & \multirow{2}{*}{\textbf{Domain}} & \textbf{Label} & \multicolumn{2}{c}{\textbf{Total Videos}} & \multicolumn{2}{c}{\textbf{Total Transitions}} & \multicolumn{3}{c}{\textbf{Transition Types}} \\
			\cmidrule(lr){4-5} \cmidrule(lr){6-7} \cmidrule(lr){8-10}
			& & \textbf{Quality} & \textbf{Count} & \textbf{Dur. (h)} & \textbf{Count} & \textbf{Dur. (s)} & \textbf{Cut} & \textbf{Normal} & \textbf{Long} \\
			\midrule
			MovieShots2 (Train) \cite{rao2020local} & Movies & Very High & 5,222 & 226.17 & 152,191 & 6,347.6 & 152,191 & 0 & 0 \\
			SportsShot (Train) \cite{mcg2024sportsshot} & Sports & Very High & 720 & 27.87 & 13,295 & 2,731.2 & 9,930 & 2,958 & 407 \\
			ClipShots (Train) \cite{tang2018fast} & Web Videos & High & 5,360 & 319.42 & 158,825 & 24,931.9 & 129,544 & 24,727 & 4,554 \\
			AutoShot (Train) \cite{zhu2023autoshot} & Short Videos & Medium & 654 & 7.36 & 5,275 & 425.1 & 4,606 & 647 & 22 \\
			\midrule
			Internal Data & Web Videos & Very High & 2,254 & 23.00 & 7,765 & 5,633.3 & 918 & 5,905 & 942 \\
			\midrule
			Synthetic Data & Web Videos & Very High & 218,993 & 959.92 & 353,198 & 604,068.9 & 40,487 & 112,158 & 200,553 \\
			\midrule
			\textbf{STD Training Data} & \textbf{Diverse} & \textbf{Mixed} & \textbf{233,203} & \textbf{1,563.74} & \textbf{690,549} & \textbf{644,138.0} & \textbf{337,676} & \textbf{146,395} & \textbf{206,478} \\
			\bottomrule
		\end{tabular}
	}
\end{table*}

To effectively optimize the \ccmethod{} for the continuous transition detection task, we establish a robust training and inference pipeline designed to overcome inherent data and memory constraints.

\noindent\textbf{Data Synthesis and Quality-Aware Sampling.} Training a VLM requires massive diverse video data with high-quality segment-level annotations. Because relying solely on legacy public datasets is insufficient due to their inherent noise and label ambiguity, we construct a scalable data engine (Figure \ref{fig:architecture}(b)) that automatically synthesizes continuous videos with random transitions and precise start/end labels. To ensure robust generalization, we aggregate this synthetic data with reformatted public datasets. Crucially, to prevent inconsistent, noisy public labels from dominating gradient updates, we implement a quality-aware mixed sampling strategy. During training, the probability of sampling a batch from a specific dataset is strictly weighted by its manually assessed quality tier, ensuring stable optimization.

\noindent\textbf{Arbitrary Video Inference.} Because the model is explicitly trained on short video clips, directly inputting arbitrarily long videos would cause out-of-distribution errors and severe memory overflow. To maintain training-inference consistency, we employ a temporal sliding-window strategy. The video stream is partitioned into overlapping temporal windows, generating local segment-level predictions for each window. Finally, we resolve redundant or conflicting transition predictions within the overlapping regions by merging them via Non-Maximum Suppression (NMS), seamlessly yielding a robust, continuous global output.

\section{Benchmark}
\label{sec:bench}

To standardize the evaluation of the newly proposed Shot Transition Detection (STD) task, we establish a comprehensive benchmark comprising customized a large-scale, high-quality test dataset and multi-dimensional evaluation metrics.

\subsection{Benchmark Construction}

\begin{table*}[t]
	\centering
	\caption{
    	\textbf{Details of proposed STD Benchmark.} Unlike SBD datasets that provide ambiguous shot boundaries, our benchmark (row 8) re-annotates existing public datasets (row 1-6) and introduces synthetic data (row 7) to evaluate different types of transitions. This equips all 5,215 videos with \textbf{segment-level} transition labels. Transition types are categorized by duration: Cut ($<0.1$s), Normal ($\leq 1$s), and Long ($>1$s).
	}
	\label{tab:benchmark_stats}
	\resizebox{\textwidth}{!}{
		\begin{tabular}{
			>{\raggedright\arraybackslash}p{5.50cm}
			>{\raggedright\arraybackslash}p{2.0cm}
			>{\centering\arraybackslash}p{2.25cm}
			>{\raggedleft\arraybackslash}p{1.25cm}
			>{\raggedleft\arraybackslash}p{1.75cm}
			>{\raggedleft\arraybackslash}p{1.25cm}
			>{\raggedleft\arraybackslash}p{1.75cm}
			>{\raggedleft\arraybackslash}p{1.5cm}
			>{\raggedleft\arraybackslash}p{1.5cm}
			>{\raggedleft\arraybackslash}p{1.25cm}
		}
			\toprule
			\multirow{2}{*}{\textbf{Dataset}} & \multirow{2}{*}{\textbf{Domain}} & \textbf{Original} & \multicolumn{2}{c}{\textbf{Total Videos}} & \multicolumn{2}{c}{\textbf{Total Transitions}} & \multicolumn{3}{c}{\textbf{Transition Types}} \\
			\cmidrule(lr){4-5} \cmidrule(lr){6-7} \cmidrule(lr){8-10}
			& & \textbf{Label} & \textbf{Count} & \textbf{Dur. (h)} & \textbf{Count} & \textbf{Dur. (s)} & \textbf{Cut} & \textbf{Normal} & \textbf{Long} \\
			\midrule
			RAI \cite{baraldi2015shot}              & TV Shows      & Point & 10    & 1.64   & 1,036   & 304.4    & 757    & 188   & 91    \\
			BBC \cite{baraldi2015scene}              & Documentaries & Point & 11    & 9.00   & 4,943   & 703.7    & 4,255  & 582   & 106   \\
			AutoShot (Test) \cite{zhu2023autoshot}  & Short Videos    & Point & 200   & 2.01   & 2,065   & 545.0    & 1,008  & 1,004 & 53    \\
			ClipShots (Test) \cite{tang2018fast} & Web Videos      & Point & 500   & 32.85  & 6,923   & 1,548.1  & 4,798  & 1,830 & 295   \\
			MovieShots2 (Test) \cite{rao2020local}      & Movies        & Point & 282   & 20.72  & 14,767  & 9,566.4  & 13,436 & 710   & 621   \\
			SportsShot (Val) \cite{mcg2024sportsshot} & Sports        & Point & 240   & 9.37   & 5,045   & 944.4    & 3,899  & 1,064 & 82    \\
			\midrule
			STD Synthesis Data           & Web Videos      & Segment  & 3,972 & 24.67  & 10,460  & 18,249.3 & 3,593  & 1,615 & 5,252 \\
			\midrule
			\textbf{STD Benchmark}  & \textbf{Diverse} & \textbf{Segment} & \textbf{5,215} & \textbf{100.26} & \textbf{45,239} & \textbf{31,861.3} & \textbf{31,746} & \textbf{6,993} & \textbf{6,500} \\
			\bottomrule
		\end{tabular}
	}
\end{table*}

Existing public datasets primarily provide noisy annotations based on isolated cut points, which are fundamentally inadequate for the segment-level requirements of the STD task. Therefore, we conducted a massive manual re-annotation process. Expert annotators meticulously reviewed the original videos and corrected the exact start and end boundaries across multiple public datasets, strictly eliminating boundary ambiguity and annotation noise.

Combined with our scalable data engine, we constructed a comprehensive evaluation benchmark encompassing 5,215 videos (approximately 100.3 hours) from diverse domains. As summarized in Table \ref{tab:benchmark_stats}, the benchmark contains 45,239 transitions. To rigorously evaluate performance across different dynamics, we categorize them by duration: abrupt cuts ($<0.1$s), normal transitions ($\leq 1$s), and long transitions ($>1$s). This accurate, segment-level ground-truth data enables robust generalization evaluation for STD models. Please refer to the Supplementary Material for more details.

\subsection{Evaluation Metrics}
\label{benchmark:eval_metr}

Traditional Shot Boundary Detection (SBD) predominantly evaluates isolated cut points independently. We argue that this point-centric evaluation is fundamentally incompatible with the STD transition detection task, which strictly requires detecting the continuous temporal span of a transition as a unified tuple. 

To this end, we propose a rigorous evaluation suite comprising segment-level and frame-level metrics. Recognizing the inherent subjective ambiguity when human annotators define the exact boundaries of gradual transitions, we introduce a temporal tolerance $\tau$ (ranging from $0.0$ to $0.5$ seconds) to expand the boundaries of ground-truth and predicted segments. We evaluate the metrics across various $\tau$ values and calculate their mean, analogous to mean Average Precision (mAP). Detailed mathematical formulations for all metrics are provided in the Supplementary Material.

\noindent\textbf{Segment-Level $F_1$.} This metric evaluates model's instance retrieval capability. A predicted segment matches a ground-truth segment if their temporal intersection is strictly positive.  Overlapping predictions are mathematically resolved via a Greedy Matching algorithm prioritizing the largest intersection duration.

\noindent\textbf{Frame-Level $F_1$.} While segment-level metrics treat all transitions equally, frame-level metrics explicitly measure the temporal coverage fidelity in the continuous time domain. Since our model outputs the start and end timestamps of a transition, we multiply these predictions by the frames per second (FPS) to obtain the exact frames for evaluation. This metric severely penalizes models that correctly detect a transition's existence but drastically misjudge its temporal span.

\noindent\textbf{Absolute Boundary Error (ABE).} To purely quantify the detection precision of the boundaries, we calculate the ABE for all successfully matched segment pairs. It measures the average absolute temporal offset (in seconds) between the predicted and ground-truth boundaries.

\noindent\textbf{Real-Time Factor (RTF).} To evaluate computational efficiency, we employ the RTF metric, defined as the ratio of the total inference time to the total duration of the processed video. An RTF strictly less than 1 indicates faster-than-real-time processing capabilities.

\section{Experiments}
\label{sec:exper}

In this section, we conduct extensive experiments to evaluate the effectiveness of the proposed \ccmethod{}. Leveraging the newly established STD benchmark, we comprehensively compare our model against three distinct paradigms of baselines: traditional heuristic approaches \cite{pyscenedetect}, specialized deep learning networks \cite{soucek2020transnet,zhu2023autoshot}, and top-tier open- and closed-source Vision-Language Models (VLMs) \cite{wang2023qwen,team2023gemini}. Empirical results consistently demonstrate the superiority of \ccmethod{} on the continuous transition detection task.

\subsection{Experimental Setup}

\noindent\textbf{Implementation Details.} To strike an optimal balance between detection performance and inference speed, we build \ccmethod{} upon the pre-trained Qwen3-VL \cite{bai2025qwen3} 4B Instruct model. As demonstrated in Table \ref{tab:experiment_comparing}, relying on larger or "Thinking" model variants severely degrades inference speed without proportional performance gains. Furthermore, because the STD task primarily demands temporal pattern recognition rather than complex reasoning, massive parameter counts are unnecessary. Conversely, the 2B Instruct model lacks the fundamental capacity for this task (consistently generating meaningless string outputs rather than valid temporal segments). Therefore, the 4B Instruct model serves as the ideal foundational backbone. 

Optical flow representations are extracted using the NeuFlow v2 \cite{zhang2024neuflowv2,zhang2024neuflow,zhang2025neuflow} architecture. The modified Vision Patch Embedding layer is initialized strictly following the zero-padding weight initialization strategy detailed in Section \ref{method:model_arch}. We optimize the model using the recently open-sourced VeOmni \cite{ma2025veomni} training framework. The network is optimized for 5,000 steps using the AdamW optimizer with a peak learning rate of $1.0 \times 10^{-5}$ and a cosine learning rate decay schedule. The per-device batch size is set to 4. All training experiments are conducted across 8 NVIDIA H100 (80GB) GPUs, yielding an effective global batch size of 32. For the sliding-window inference, we strictly adhere to a window size of $10$ seconds with a temporal stride of $9$ seconds. Powered by FFmpeg \cite{tomar2006converting}, our engine supports the generation of 59 distinct transition effects (see Appendix S2 for details). For the quality-aware mixed sampling strategy, the sampling probabilities for Very High, High, and Medium tiers are $0.7$, $0.2$, and $0.1$, respectively.

\noindent\textbf{Baselines and Fair Comparison.} We benchmark \ccmethod{} against three distinct categories of representative methods: (1) \textit{Heuristic algorithms}, represented by the widely adopted PySceneDetect \cite{pyscenedetect}; (2) \textit{Specialized deep learning networks}, including the current state-of-the-art SBD models TransNetV2 \cite{soucek2020transnet} and AutoShot \cite{zhu2023autoshot}; and (3) \textit{General-purpose foundational VLMs}, specifically evaluating the Qwen3-VL \cite{bai2025qwen3} and Gemini \cite{team2023gemini} series. To guarantee a strictly fair and rigorous comparison, the predictions generated by SBD baselines are systematically converted to transition segments through simple temporal subtraction. All comprehensive evaluations are standardized on our newly established STD benchmark. To balance fairness and evaluation efficiency, we set the video FPS to 5 when evaluating general VLMs.

\subsection{Quantitative Experiments}

\begin{table*}[t]
	\centering
	\caption{
    	\textbf{Quantitative comparison on the STD Benchmark.} We report both mean segment-level and frame-level metrics (Precision, Recall, and $F_1$) across the defined temporal tolerance values to comprehensively evaluate the detection performance. Performance is measured by the Absolute Boundary Error (ABE) in seconds. Inference efficiency is reported via the Real-Time Factor (RTF). The evaluations are conducted on both public and synthetic datasets. The best results are highlighted in \textbf{bold}, and the second-best are \underline{underlined}. Our proposed \ccmethod{} outperforms state-of-the-art SBD methods and general VLMs. P means Precision and R means Recall.
	}
	\label{tab:experiment_comparing}
	\resizebox{\textwidth}{!}{
    \begin{tabular}{
	    >{\raggedright\arraybackslash}p{4.0cm}
	    >{\centering\arraybackslash}p{1.0cm}
	    >{\centering\arraybackslash}p{1.0cm}
	    >{\centering\arraybackslash}p{1.0cm}
	    >{\centering\arraybackslash}p{1.0cm}
	    >{\centering\arraybackslash}p{1.0cm}
	    >{\centering\arraybackslash}p{1.0cm}
	    >{\centering\arraybackslash}p{1.0cm}
	    >{\centering\arraybackslash}p{1.0cm}
	    >{\centering\arraybackslash}p{1.0cm}
	    >{\centering\arraybackslash}p{1.0cm}
	    >{\centering\arraybackslash}p{1.0cm}
	    >{\centering\arraybackslash}p{1.0cm}
	    >{\centering\arraybackslash}p{1.0cm}
	    >{\centering\arraybackslash}p{1.0cm}
	    >{\centering\arraybackslash}p{1.0cm}
	}
        \toprule
        & \multicolumn{7}{c}{\textbf{Public Data}} & \multicolumn{7}{c}{\textbf{Synthetic Data}} & \\
        \cmidrule(lr){2-8} \cmidrule(lr){9-15}
        & \multicolumn{3}{c}{Segment} & \multicolumn{3}{c}{Frame} & & \multicolumn{3}{c}{Segment} & \multicolumn{3}{c}{Frame} & & \\
        \cmidrule(lr){2-4} \cmidrule(lr){5-7} \cmidrule(lr){9-11} \cmidrule(lr){12-14}
        Methods & P & R & F1 & P & R & F1 & ABE & P & R & F1 & P & R & F1 & ABE & RTF \\
        \midrule
        \multicolumn{10}{l}{PySceneDetect \cite{pyscenedetect}} \\
        \midrule
        \quad - Adaptive & 0.656 & 0.716 & 0.684 & 0.645 & 0.372 & 0.457 & 1.93 & \textbf{0.924} & 0.172 & 0.290 & 0.922 & 0.036 & 0.069 & \underline{0.28} & 0.09 \\
        \quad - Content & 0.627 & 0.759 & 0.686 & 0.603 & 0.383 & 0.453 & \underline{1.08} & \underline{0.909} & 0.128 & 0.225 & \textbf{0.968} & 0.029 & 0.056 & 0.61 & 0.09 \\
        \quad - Hash & 0.566 & 0.778 & 0.654 & 0.549 & 0.416 & 0.457 & 2.11 & 0.880 & 0.118 & 0.208 & 0.940 & 0.027 & 0.052 & 0.85 & 0.09 \\
        \quad - Hist & 0.452 & 0.741 & 0.559 & 0.419 & 0.398 & 0.394 & 1.19 & 0.577 & 0.379 & 0.455 & 0.762 & 0.117 & 0.197 & 1.04 & 0.09 \\
        \quad - Threshold & 0.364 & 0.031 & 0.057 & 0.306 & 0.014 & 0.027 & 3.08 & 0.773 & 0.039 & 0.074 & 0.749 & 0.008 & 0.016 & 1.46 & 0.09 \\
        \midrule
        TransNetV2 \cite{soucek2020transnet} & \underline{0.727} & 0.780 & \underline{0.752} & \textbf{0.731} & 0.427 & 0.528 & 1.87 & 0.275 & 0.149 & 0.194 & 0.417 & 0.034 & 0.063 & 0.61 & \underline{0.07} \\
        \midrule
        AutoShot \cite{zhu2023autoshot} & 0.707 & \underline{0.804} & 0.751 & \underline{0.709} & \underline{0.440} & \underline{0.532} & 1.78 & 0.379 & 0.248 & 0.299 & 0.530 & 0.058 & 0.102 & 0.51 & \textbf{0.03} \\
        \midrule
        \multicolumn{10}{l}{Gemini Series \cite{team2023gemini}} \\
        \midrule
        \quad - 2.5 Pro & 0.558 & 0.527 & 0.542 & 0.453 & 0.361 & 0.401 & 3.62 & 0.338 & \underline{0.851} & 0.465 & 0.638 & \underline{0.760} & \underline{0.686} & 0.82 & 0.81 \\
        \quad - 3 Pro & 0.527 & 0.573 & 0.549 & 0.469 & 0.343 & 0.393 & 2.18 & 0.479 & 0.768 & 0.588 & 0.711 & 0.482 & 0.568 & 0.88 & 1.32 \\
        \midrule
        \multicolumn{10}{l}{Qwen3-VL Instruct Series \cite{bai2025qwen3}} \\
        \midrule
        \quad - 4B & 0.235 & 0.088 & 0.124 & 0.134 & 0.174 & 0.148 & 55.00 & 0.717 & 0.306 & 0.428 & 0.458 & 0.618 & 0.525 & 8.88 & 0.31 \\
        \quad - 8B & 0.222 & 0.297 & 0.246 & 0.132 & 0.307 & 0.184 & 34.39 & 0.586 & 0.597 & 0.591 & 0.741 & 0.204 & 0.315 & 1.60 & 0.34 \\
        \quad - 32B & 0.309 & 0.473 & 0.370 & 0.214 & 0.279 & 0.241 & 3.96 & 0.895 & 0.623 & \underline{0.735} & 0.911 & 0.361 & 0.516 & 1.35 & 0.98 \\
        \quad - 30B-A3B(MoE) & 0.218 & 0.300 & 0.242 & 0.171 & 0.222 & 0.192 & 9.61 & 0.806 & 0.593 & 0.683 & 0.749 & 0.355 & 0.482 & 1.86 & 0.35 \\
        \midrule
        \multicolumn{10}{l}{Qwen3-VL Thinking Series \cite{bai2025qwen3}} \\
        \midrule
        \quad - 4B & 0.449 & 0.079 & 0.134 & 0.265 & 0.051 & 0.086 & 1.50 & 0.839 & 0.218 & 0.346 & 0.748 & 0.087 & 0.156 & 1.68 & 1.03 \\ 
        \quad - 8B & 0.450 & 0.144 & 0.217 & 0.297 & 0.094 & 0.143 & 4.25 & 0.854 & 0.389 & 0.534 & 0.923 & 0.147 & 0.252 & 1.29 & 1.31 \\ 
        \quad - 32B & 0.403 & 0.260 & 0.315 & 0.308 & 0.160 & 0.209 & 1.09 & 0.900 & 0.608 & 0.726 & 0.943 & 0.323 & 0.479 & 1.16 & 3.30 \\ 
        \quad - 30B-A3B(MoE) & 0.374 & 0.217 & 0.275 & 0.291 & 0.127 & 0.175 & \textbf{0.98} & 0.890 & 0.610 & 0.724 & 0.915 & 0.331 & 0.485 & 1.18 & 0.92 \\
        \midrule
        \textbf{TransVLM(Ours)} & \textbf{0.762} & \textbf{0.806} & \textbf{0.783} & 0.574 & \textbf{0.562} & \textbf{0.568} & 1.58 & 0.908 & \textbf{0.882} & \textbf{0.895} & \underline{0.946} & \textbf{0.930} & \textbf{0.938} & \textbf{0.11} & 0.50 \\ 
		\bottomrule
    \end{tabular}}
\end{table*}

The comprehensive quantitative results evaluating \ccmethod{} against SBD approaches and top-tier foundational VLMs are summarized in Table \ref{tab:experiment_comparing}. We follow the official calling guidance for each method. Overall, our proposed framework establishes a new state-of-the-art across both public and synthetic datasets, consistently dominating the primary $F_1$ metrics. See Appendix S5 for details.

\noindent\textbf{Performance on Public Data.} On the public datasets, \ccmethod{} significantly outperforms all baselines in accurately detecting transition instances and their temporal coverage. Specifically, it achieves the highest segment-level $F_1$ ($78.3\%$) and frame-level $F_1$ ($56.8\%$), surpassing highly specialized deep learning models such as AutoShot ($75.1\%$ and $53.2\%$, respectively). Our Absolute Boundary Error (ABE) of $1.58$s is highly competitive. This marginal variance from the absolute lowest score is largely attributable to the inherent limitations of public training datasets, which severely suffer from noisy, subjective, and ambiguous human annotations (detailed examples and discussions are provided in the Appendix S4). Consequently, these subjective inconsistencies inevitably cap the measurable upper bound of precise boundary detection on public test sets.

Notably, on public data, traditional SBD methods significantly outperform general VLMs. Because transitions in these public datasets are predominantly abrupt cuts, this observation corroborates our hypothesis regarding VLM deficiencies: VLMs possess insufficient perception of fine-grained inter-shot motion changes, and their reliance on sparse, low-frame-rate inputs severely restricts their ability to detect instantaneous abrupt cuts. 

\noindent\textbf{Performance on Synthetic Data.} Unlike public data, the synthetic data purposefully contains a massive proportion of challenging scenarios, including subtle cuts, complex gradual transitions, and prolonged transitions spanning multiple seconds. Here, \ccmethod{} demonstrates absolute dominance, achieving an unprecedented segment-level $F_1$ of $89.5\%$, a frame-level $F_1$ of $93.8\%$, and an astonishingly low mABE of just $0.11$s. 

When confronted with these diverse dynamics, SBD methods experience a catastrophic performance drop. For instance, AutoShot's frame-level $F_1$ plummets to $10.2\%$, exposing its fundamental bias towards simple, abrupt cuts. Interestingly, on the synthetic data, general VLMs generally outperform SBD methods. This further validates our critique of SBD limitations: SBD methods focus heavily on searching for ambiguous isolated cut points rather than capturing the intrinsic spatiotemporal patterns of complete transition segments. However, while general-purpose VLMs exhibit some inherent cognitive capacity to perceive gradual transitions, they drastically fail to accurately detect abrupt cuts, resulting in elevated mABE scores. \ccmethod{} is the only framework that seamlessly and robustly unifies the detection of all transition types.

\noindent\textbf{Inference Efficiency.} Beyond state-of-the-art detection performance, \ccmethod{} maintains highly practical inference efficiency. As reported in Table \ref{tab:experiment_comparing}, our model achieves a Real-Time Factor (RTF) of $0.50$. An RTF strictly less than $1.0$ signifies faster-than-real-time processing capabilities (i.e., processing one second of video takes only $0.5$ seconds). While lightweight heuristic algorithms and heavily down-sampled CNNs (e.g., AutoShot at RTF $0.03$) naturally execute faster due to their simplistic architectures, their fundamental inability to parse complex, gradual transitions renders them inadequate for modern high-quality video generation pipelines. \ccmethod{} strikes the optimal trade-off, delivering unprecedented temporal detection precision without compromising real-world deployment viability.

\subsection{Ablation Studies}

\begin{table*}[t]
	\centering
	\caption{
    	\textbf{Ablation Study of \ccmethod{}.} The ablations include: (1) \textit{Training}: identifying the importance of our training process; (2) \textit{Data Composition}: isolating the impact of our data engine by removing either public or synthetic data; (3) \textit{Motion Prior}: removing the optical flow input; (4) \textit{Fusion Strategy}: substituting our feature fusion strategy by inputting color and optical flow frames separately (\textit{w/o feature fusion}), which drastically increases RTF; and (5) \textit{Initialization}: replacing our zero-padding initialization with a naive channel duplication (\textit{duplicate weight}). The results demonstrate that our full \ccmethod{} configuration achieves the optimal balance between detection performance (highest overall F1) and inference efficiency (RTF). Best results are in \textbf{bold} and second-best are \underline{underlined}. P means Precision and R means Recall.
	}
	\label{tab:experiment_ablation}
	\resizebox{\textwidth}{!}{
    \begin{tabular}{
	    >{\raggedright\arraybackslash}p{2.9cm}
	    >{\centering\arraybackslash}p{1.0cm}
	    >{\centering\arraybackslash}p{1.0cm}
	    >{\centering\arraybackslash}p{1.0cm}
	    >{\centering\arraybackslash}p{1.0cm}
	    >{\centering\arraybackslash}p{1.0cm}
	    >{\centering\arraybackslash}p{1.0cm}
	    >{\centering\arraybackslash}p{1.0cm}
	    >{\centering\arraybackslash}p{1.0cm}
	    >{\centering\arraybackslash}p{1.0cm}
	    >{\centering\arraybackslash}p{1.0cm}
	    >{\centering\arraybackslash}p{1.0cm}
	    >{\centering\arraybackslash}p{1.0cm}
	    >{\centering\arraybackslash}p{1.0cm}
	    >{\centering\arraybackslash}p{1.0cm}
	    >{\centering\arraybackslash}p{1.0cm}
	}
        \toprule
        & \multicolumn{7}{c}{\textbf{Real-world Data}} & \multicolumn{7}{c}{\textbf{Synthetic Data}} & \\
        \cmidrule(lr){2-8} \cmidrule(lr){9-15}
        & \multicolumn{3}{c}{Segment} & \multicolumn{3}{c}{Frame} & & \multicolumn{3}{c}{Segment} & \multicolumn{3}{c}{Frame} & & \\
        \cmidrule(lr){2-4} \cmidrule(lr){5-7} \cmidrule(lr){9-11} \cmidrule(lr){12-14}
        Methods & P & R & F1 & P & R & F1 & ABE & P & R & F1 & P & R & F1 & ABE & RTF \\
        \midrule
        w/o training & 0.545 & 0.448 & 0.487 & 0.265 & 0.501 & 0.343 & 7.630 & 0.809 & 0.478 & 0.599 & 0.701 & 0.304 & 0.424 & 2.085 & 0.76 \\
        w/o real-world data & 0.756 & 0.287 & 0.416 & 0.330 & 0.223 & 0.264 & 1.700 & \underline{0.962} & \underline{0.894} & \underline{0.927} & \underline{0.957} & \underline{0.926} & \textbf{0.941} & \underline{0.114} & 0.51 \\
        w/o synthetic data & 0.755 & 0.706 & 0.730 & \underline{0.587} & 0.483 & 0.530 & 1.849 & \textbf{0.972} & 0.572 & 0.720 & 0.857 & 0.656 & 0.743 & 0.455 & \textbf{0.50} \\
        w/o optical flow & \textbf{0.793} & 0.617 & 0.694 & 0.574 & 0.435 & 0.495 & \textbf{1.299} & 0.946 & 0.783 & 0.857 & 0.954 & 0.838 & 0.892 & 0.136 & 0.69 \\
        w/o feature fusion & 0.700 & \underline{0.782} & \underline{0.739} & \textbf{0.609} & \underline{0.518} & \underline{0.557} & 1.921 & 0.931 & \textbf{0.931} & \textbf{0.931} & \textbf{0.957} & 0.925 & \underline{0.941} & 0.120 & 1.29 \\
        duplicate weight & \underline{0.767} & 0.575 & 0.655 & 0.573 & 0.419 & 0.484 & 1.786 & 0.958 & 0.830 & 0.889 & 0.949 & 0.909 & 0.928 & 0.143 & 0.69 \\
        \midrule
        \textbf{TransVLM(Ours)} & 0.762 & \textbf{0.806} & \textbf{0.783} & 0.574 & \textbf{0.562} & \textbf{0.568} & \underline{1.576} & 0.908 & 0.882 & 0.895 & 0.946 & \textbf{0.930} & 0.938 & \textbf{0.113} & \underline{0.50} \\
		\bottomrule
    \end{tabular}}
\end{table*}

To validate the contribution of each component in \ccmethod{}, we conduct comprehensive ablation studies. The quantitative results across both public and synthetic datasets are detailed in Table \ref{tab:experiment_ablation}. See Appendix S6 for details.

\noindent\textbf{Ablation on the Training Process.} In order to identify the importance of our training process, we evaluate the pre-trained model directly using our arbitrary video inference strategy as a baseline. The results show that without subsequent fine-tuning, the general VLM performs poorly across all dimensions (e.g., yielding only $48.7\%$ segment-level $F_1$ on public data). However, compared to the direct application of FPS=5 shown in Table \ref{tab:experiment_comparing}, utilizing our sliding-window inference strategy yields noticeable improvements across most metrics. This underscores that while off-the-shelf VLMs lack the fundamental zero-shot capability for precise transition detection, our inference strategy is structurally beneficial. 

\noindent\textbf{Ablation on the Data Composition.} We first evaluate the necessity of our fine-tuning paradigm and the constructed datasets by isolating the effects of our mixed-data training strategy. Training exclusively on synthetic data (\textit{w/o public data}) achieves excellent results on the synthetic test set but suffers a severe domain shift when evaluated on public videos, plummeting to $41.6\%$ segment-level $F_1$. Conversely, training solely on public datasets (\textit{w/o synthetic data}) yields suboptimal generalization on complex synthetic transitions. This stark contrast highlights a significant domain gap in transition distributions (as previously foreshadowed in Table \ref{tab:benchmark_stats}). By implementing a quality-aware mixed training strategy, \ccmethod{} successfully bridges this gap, leveraging the vast diversity of the synthetic data while robustly anchoring its spatial understanding to public visual distributions, achieving an optimal segment-level $F_1$ of $78.3\%$ on public data.

\noindent\textbf{Ablation on the Motion Prior.} Gradual transitions are inherently defined by inter-frame pixel dynamics. When the optical flow input is entirely removed (\textit{w/o optical flow}), the model is forced to rely solely on color spatial semantics. This causes a significant performance degradation, dropping the public segment-level $F_1$ from $78.3\%$ to $69.4\%$, and the synthetic frame-level $F_1$ from $93.8\%$ to $89.2\%$. This validates our core hypothesis: explicitly injecting optical flow effectively compensates for VLM's inherent insensitivity to low-level temporal motion cues.

\noindent\textbf{Ablation on the Fusion Strategy.} To demonstrate the efficiency of our simple feature-fusion mechanism, we evaluate an alternative (\textit{w/o feature fusion}), where the color and optical flow frames are processed as two separate visual inputs via different vision encoders to the VLM. While this uncompressed paradigm achieves highly competitive performance (e.g., $94.1\%$ frame-level $F_1$ on synthetic data), it brutally inflates the sequence length of visual tokens. Consequently, the self-attention computational overhead skyrockets, deteriorating the Real-Time Factor (RTF) from a highly efficient $0.50$ to a sluggish $1.29$, rendering it entirely unsuitable for real-time video processing. Our simple feature-fusion modification perfectly balances temporal sensitivity with strict inference efficiency. 

\noindent\textbf{Ablation on the Zero-Padding Initialization.} Finally, we ablate our mathematically stable zero-padding initialization strategy by substituting it with a naive channel duplication method (\textit{duplicate weight}), where the pre-trained color weights are copied and scaled for the optical flow channels. As evidenced by the sharp drop in public segment-level $F_1$ ($78.3\% \rightarrow 65.5\%$), forcefully introducing large, uncalibrated motion gradients during the initial fine-tuning epochs severely destabilizes the optimization process. It causes catastrophic forgetting of the VLM's pre-trained spatial representations. Our zero-padding strategy effectively neutralizes this risk, ensuring stable convergence.

\section{Conclusion}
\label{sec:conclusion}

In this work, we address the fundamental limitations of traditional Shot Boundary Detection (SBD) by formulating the Shot Transition Detection (STD) task. We propose \ccmethod{}, a Vision-Language Model (VLM) framework that explicitly integrates optical flow as a motion prior alongside standard color frames via a parameter-efficient feature-fusion strategy. This design significantly enhances the model's temporal awareness for detecting diverse transitions without incurring any additional visual token overhead. Furthermore, to overcome the severe class imbalance and annotation noise in public data, we develop a scalable data engine for robust VLM optimization and establish a comprehensive STD benchmark rigorously equipped with multi-dimensional metrics with temporal tolerance. Extensive experiments demonstrate that \ccmethod{} achieves superior overall performance, substantially outperforming traditional heuristic methods, specialized networks, and top-tier VLMs.

{\small
\bibliographystyle{plainnat}
\bibliography{main}
}

\clearpage
\appendix
\section{Appendix}
\subsection{Extended Details of Motivation}
\label{sec:appendix_intro_details}

\noindent\textbf{Detailed Transition Definitions and SBD Limitations.} A video shot is defined as a continuous sequence of frames captured seamlessly by a single, uninterrupted camera. It serves as the fundamental, structurally indivisible unit of a video, wherein all intra-shot frames share consistent visual content and coherent motion. Conversely, a transition is the temporal sequence bridging two adjacent shots. Its duration can vary drastically, ranging from an instantaneous abrupt cut to a prolonged gradual progression. When traditional SBD methods \cite{abdulhussain2018methods, kar2024video, pyscenedetect, zabih1995feature, xu2016shot, wu2019two, tong2015cnn, soucek2019transnet, soucek2020transnet, zhu2023autoshot} process gradual transitions (e.g., dissolves, fades, and wipes), they often fail to determine precise boundaries, resulting in extracted shots contaminated by dirty transitional frames. Furthermore, these methods frequently miss subtle cuts (cuts with minor content changes), long gradual transitions, and special effect transitions. By explicitly detecting these continuous shot transition segments in our formulated STD task, pure and clean shots can be efficiently extracted through simple temporal subtraction.

\noindent\textbf{Benchmark Class Imbalance and Data Engine Synthesis.} During the construction of our benchmark, we meticulously re-annotated widely used public data (e.g., RAI \cite{baraldi2015shot}, BBC \cite{baraldi2015scene}, AutoShot \cite{zhu2023autoshot}, and MovieShots \cite{rao2020local}). However, through random sampling analysis, we observed that these public data suffer from severe class imbalance. They are heavily dominated by normal cuts ($\sim$80\%), with only a marginal presence of gradual transitions ($\sim$15\%) and exceptionally few long transitions ($\sim$5\%). This skewed distribution fundamentally hinders a comprehensive evaluation. Leveraging our scalable data engine, we generated diverse synthetic data comprising normal and subtle cuts ($\sim$30\%), gradual transitions ($\sim$20\%), and prolonged gradual or special effect transitions ($\sim$50\%), thereby enabling a rigorous evaluation across all diverse transition dynamics.

\noindent\textbf{Multi-dimensional Evaluation Protocol.} Recognizing that traditional SBD metrics provide an incomplete assessment for the STD task, we establish a multi-dimensional evaluation protocol. Specifically, we introduce segment-level $F_1$ to evaluate the overall performance of transition detection, frame-level $F_1$ to measure overall temporal coverage, mean Absolute Boundary Error (ABE) to analyze boundary offsets, and Real-Time Rate (RTR) to evaluate inference speed. To account for the inherent ambiguity in human annotation, our metrics explicitly incorporate a temporal tolerance applied exclusively to the segment annotations and model predictions, ensuring a robust and equitable benchmarking process.

\noindent\textbf{Analysis of Visual Tokens for Optical Flow.} To validate the efficacy of introducing explicit motion priors, we visualize the feature maps of the visual tokens extracted immediately after the vision model and prior to the language model. For a given video segment, the optical flow of the current frame is computed relative to its preceding frame. To generate the feature maps, we apply a max-pooling operation along the feature channel dimension of the visual tokens. The feature tensors are then visualized using a sequential colormap, where lower values are rendered in darker color (approaching black) and higher activations in brighter color (approaching yellow). It is worth noting that the pre-trained vision model of Qwen3-VL intrinsically employs a temporal downsampling stride of 2; thus, every two consecutive raw video frames are compressed into a single temporal visual token representation.

\begin{figure}[htbp]
    \centering
    \includegraphics[width=0.9\textwidth]{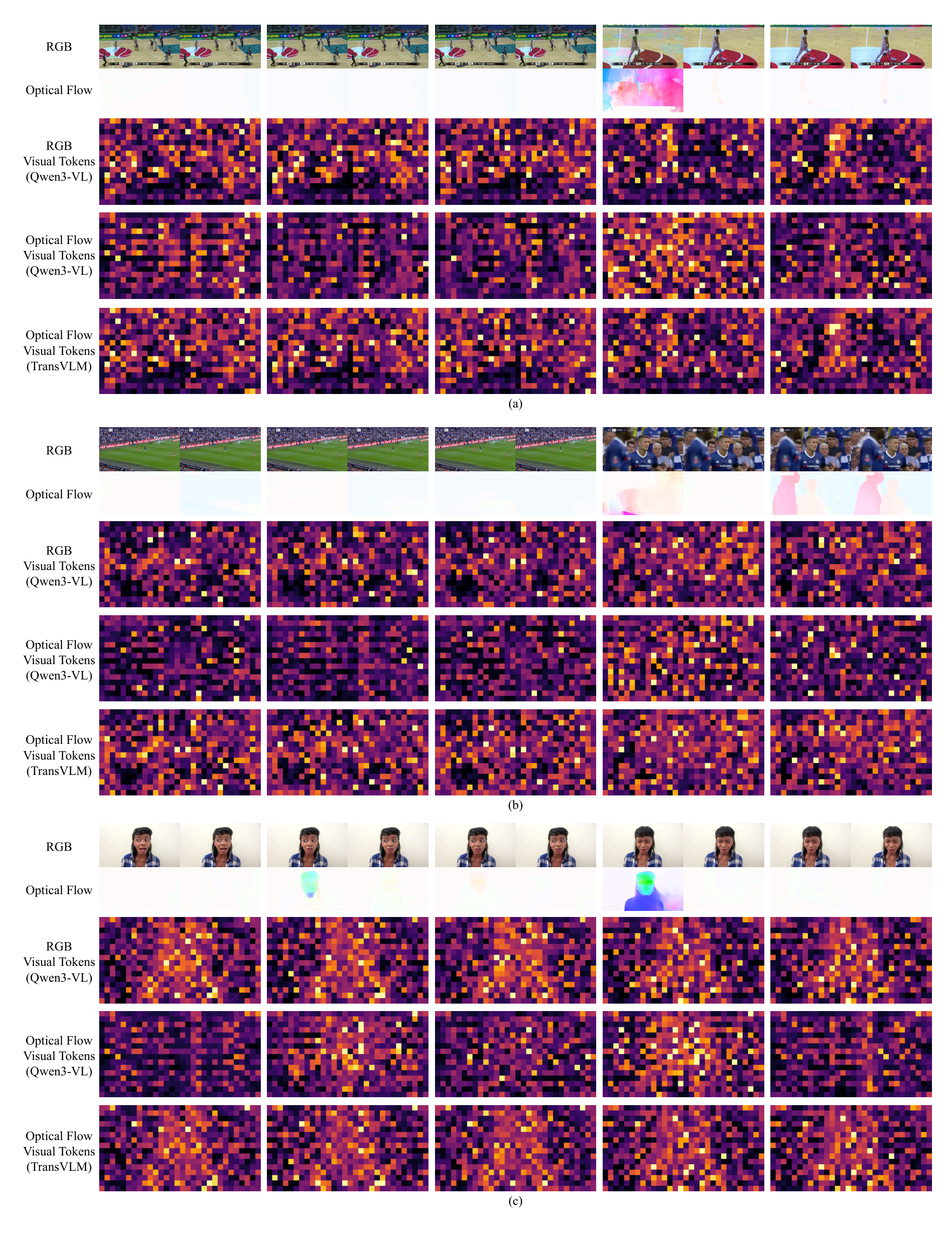}
    \caption{\textbf{Feature map visualizations of visual tokens extracted by vision model (Part A).} The feature maps are obtained via channel-wise max pooling, with brighter yellow indicating higher magnitudes. While \ccmethod{}'s fused tokens generally resemble RGB representations due to zero-padding initialization, they exhibit distinctly sharper temporal contrast at transition boundaries, proving the enhanced sensitivity brought by the optical flow motion prior.}
    \label{fig:appendix_visual_tokens_a}
\end{figure}

\begin{figure}[htbp]
    \centering
    \includegraphics[width=0.8\textwidth]{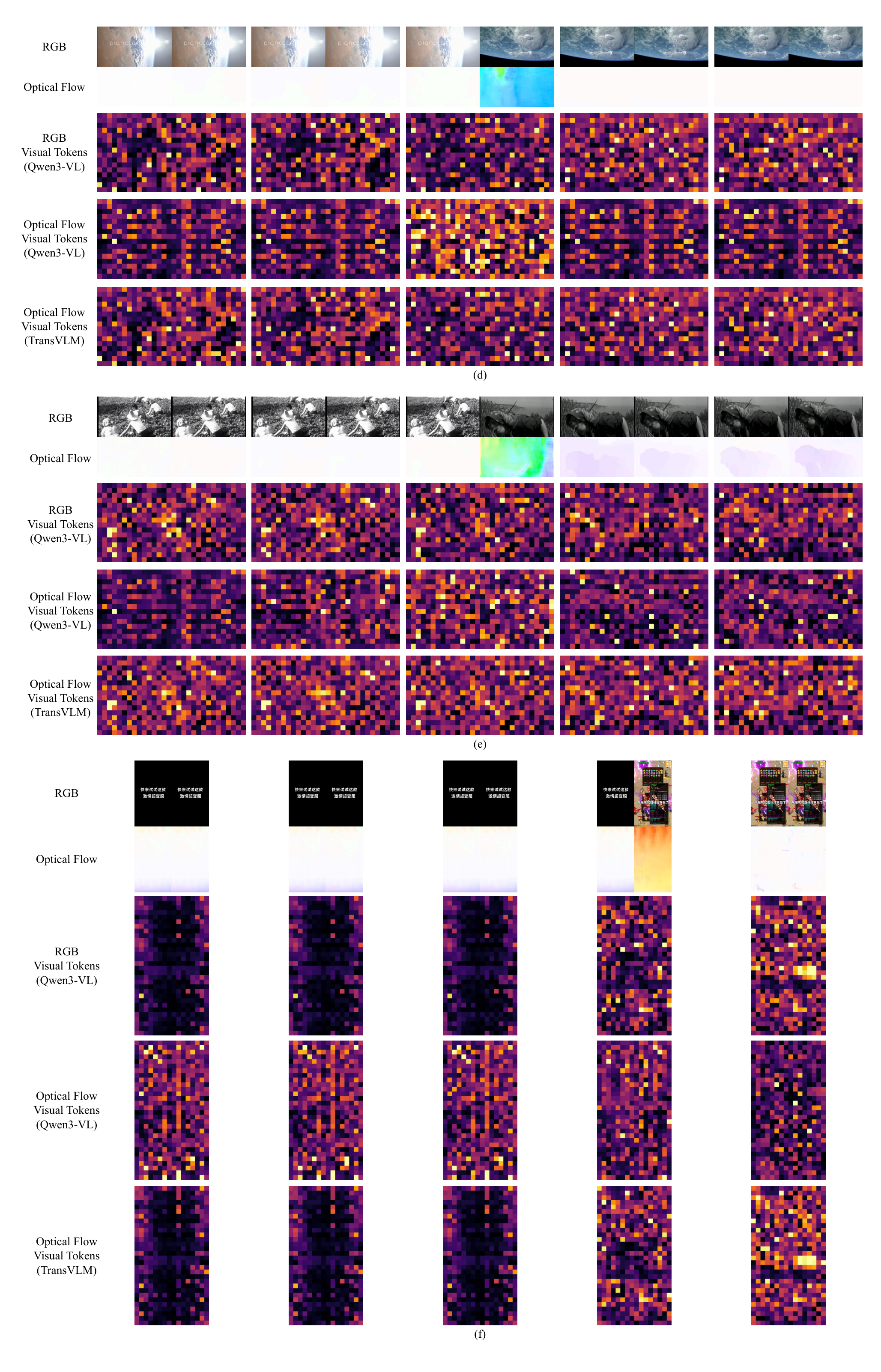}
    \caption{\textbf{Feature map visualizations of visual tokens extracted by vision model (Part B).} The feature maps are obtained via channel-wise max pooling, with brighter yellow indicating higher magnitudes. While \ccmethod{}'s fused tokens generally resemble RGB representations due to zero-padding initialization, they exhibit distinctly sharper temporal contrast at transition boundaries, proving the enhanced sensitivity brought by the optical flow motion prior.}
    \label{fig:appendix_visual_tokens_b}
\end{figure}

As illustrated in Figure \ref{fig:appendix_visual_tokens_a} and \ref{fig:appendix_visual_tokens_b}, for abrupt shot cut---encompassing both standard hard cuts and highly challenging subtle cuts---pure optical flow modalities distinctly capture the inter-frame motions. Consequently, the visual tokens derived from optical flow exhibit significantly stronger response at cut frames when compared to the tokenized representations of pure RGB frames.

In our proposed \ccmethod{}, the RGB and optical flow modalities are fused at the data level when fed to the vision model. Because we employ a zero-padding initialization strategy to guarantee training stability and preserve pre-trained knowledge, the macroscopic pattern of \ccmethod{}'s visual tokens visually resembles that of the pure RGB tokens. However, upon closer inspection (e.g., Case (c) in Figure \ref{fig:appendix_visual_tokens_a}), the fused feature maps at the cut frames demonstrate a sharper contrast against adjacent frames than their RGB-only counterparts. This enhancement provides compelling evidence that the injected motion prior could augment the VLM's sensitivity to fine-grained inter-frame motions without corrupting its foundational spatial understanding.

\subsection{Extended Details of Benchmark and Data Engine}
\label{sec:supp_bench}

\subsubsection{Detailed Metric Formulations}

For both segment-level and frame-level evaluations in the STD task, we compute Precision ($P$), Recall ($R$), and the $F_1$-score using the standard formulations:
\begin{equation}
	P = \frac{TP}{TP + FP}, \quad R = \frac{TP}{TP + FN}, \quad F_1 = \frac{2 \times P \times R}{P + R}.
\end{equation}

For \textbf{Segment-Level metrics}, True Positives (TP) correspond to the total count of successfully matched predicted segments; False Positives (FP) represent the count of unmatched predictions; and False Negatives (FN) denote the count of ground-truths that fail to match with any prediction.

For \textbf{Frame-Level metrics}, TP is defined as the total number of frames within the intersections across all successfully matched prediction and ground-truth segment pairs. FP is the sum of the frames across all predicted segments minus the TP frames. Symmetrically, FN is the sum of the frames across all ground-truth segments minus the TP frames. 

To quantify the precision of the boundary localization, the mean Absolute Boundary Error (ABE) is formally calculated as:
\begin{equation}
	ABE = \frac{1}{2 |TP|} \sum_{(P,G) \in TP} (|g_s - p_s| + |g_e - p_e|),
\end{equation}
where $(p_s, p_e)$ and $(g_s, g_e)$ represent the start and end timestamps of the predicted boundaries and the ground-truth boundaries, respectively.

To critically evaluate the computational efficiency of the models, we report the Real-Time Factor (RTF). It is mathematically defined as the ratio of the total inference time required by the model to the total temporal duration of the processed video:
\begin{equation}
	RTF = \frac{T_{\text{inference}}}{T_{\text{video}}},
\end{equation}
where $T_{\text{inference}}$ denotes the absolute processing time (in seconds) and $T_{\text{video}}$ represents the actual duration of the input video sequence (in seconds). An RTF strictly less than $1.0$ signifies that the model achieves faster-than-real-time processing, which is a highly desirable property for deploying the transition detection task in practical, large-scale video pipelines.

\subsubsection{Benchmark Dataset Distribution}

To ensure robust generalization evaluation, the 5,215 videos constituting our STD benchmark are carefully sourced from highly diverse domains. Based on the total temporal duration, the benchmark comprises television shows (RAI \cite{baraldi2015shot}, $1.6\%$), documentaries (BBC \cite{baraldi2015scene}, $9.0\%$), short mobile videos (AutoShot \cite{zhu2023autoshot}, $2.0\%$), web videos (ClipShots \cite{tang2018fast}, $32.8\%$), movies (MovieShots2 \cite{rao2020local}, $20.7\%$), sports broadcasting (SportsShot \cite{mcg2024sportsshot}, $9.3\%$), and our synthesized dataset (STD-Synth, $24.6\%$). The synthesized dataset systematically injects varying transition patterns, including abrupt cuts, normal fades, and prolonged transitions, into continuous pure shots to systematically test model robustness.

\subsubsection{Strict Re-annotation Quality Control}

Because existing public SBD datasets primarily provide noisy annotations centered on isolated cut points, they are fundamentally inadequate for the segment-level requirements of the continuous transition detection task. To meet these rigorous demands, we conducted a massive manual re-annotation process. Expert annotators meticulously reviewed the original videos and corrected the exact start and end boundaries of the transitions across all utilized public datasets (RAI \cite{rao2020local}, BBC \cite{baraldi2015scene}, AutoShot \cite{zhu2023autoshot}, ClipShots \cite{tang2018fast}, MovieShots2 \cite{rao2020local}, and SportsShot \cite{mcg2024sportsshot}). This strict quality control process establishes a highly accurate, segment-level ground-truth reference tailored specifically for comprehensively evaluating STD models.

\clearpage

\subsubsection{Data Engine Based on FFmpeg \cite{tomar2006converting}} 

To construct a robust and highly scalable training pipeline for the Shot Transition Detection (STD) task, we developed an automated data synthesis engine powered by FFmpeg. This engine seamlessly concatenates discrete pure video shots while injecting diverse, randomized temporal transition dynamics, overcoming the severe class imbalance and annotation noise inherent in legacy public data.

\noindent\textbf{Transition Synthesis Logic.} Given a sequential pool of pure video shots, the data engine automatically synthesizes continuous video streams through a rigorous randomized sampling mechanism. For any two adjacent shots, the engine defines a maximum allowable transition duration, denoted as $\mathcal{T}_{max}$. This upper bound is dynamically constrained by the actual temporal lengths of the incoming and outgoing adjacent shots to strictly prevent temporal boundary overflow. 

Subsequently, the engine explicitly samples a continuous transition duration $d \sim \mathcal{U}(0, \mathcal{T}_{max})$ and randomly uniformly selects a specific transition effect from a comprehensive pool of 59 distinct transition types. Crucially, if the abruptly transitioning \textit{``cut''} type is selected, the transition duration $d$ is strictly overridden to $0$. Once the parameters are sampled, the engine utilizes FFmpeg's advanced rendering pipeline to synthesize the fused video clip. Concurrently, it accurately calculates the precise start and end timestamps of the injected transition segment, automatically exporting these segment-level temporal tuples into a structured JSON label file. This paradigm inherently guarantees the generation of $100\%$ noise-free, accurate segment-level ground truth required for robust Vision-Language Model (VLM) training.

\noindent\textbf{Supported Transition Types.} To ensure the trained model robustly generalizes across diverse temporal and spatial transition patterns, our data engine supports the synthesis of 59 distinct transition effects. The exhaustive list, along with their spatiotemporal definitions, is provided below:

\begin{enumerate}
    \item \textbf{cut}: An instantaneous, abrupt switch between shots with exactly zero temporal duration.
    \item \textbf{fade}: A standard crossfade where the outgoing shot smoothly decreases in opacity while the incoming shot increases.
    \item \textbf{wipeleft}: A linear spatial wipe transitioning across the frame from right to left.
    \item \textbf{wiperight}: A linear spatial wipe transitioning from left to right.
    \item \textbf{wipeup}: A vertical linear wipe progressing from the bottom edge to the top edge.
    \item \textbf{wipedown}: A vertical linear wipe progressing from the top edge to the bottom edge.
    \item \textbf{slideleft}: Translates the incoming shot globally over the outgoing shot from right to left.
    \item \textbf{slideright}: Translates the incoming shot horizontally from left to right.
    \item \textbf{slideup}: Translates the incoming shot vertically upwards from the bottom edge.
    \item \textbf{slidedown}: Translates the incoming shot vertically downwards from the top edge.
    \item \textbf{circlecrop}: Reveals the incoming shot through an expanding circular spatial mask originating from the center.
    \item \textbf{rectcrop}: Reveals the incoming shot through a symmetrically expanding rectangular mask.
    \item \textbf{distance}: Blends pixels temporally based on spatial distance calculations between the two frames.
    \item \textbf{fadeblack}: Fades the outgoing shot completely to a solid black frame before fading into the incoming shot.
    \item \textbf{fadewhite}: Fades the outgoing shot completely to a solid white frame before fading into the incoming shot.
    \item \textbf{radial}: A circular, clock-like rotational sweep mask revealing the incoming shot.
    \item \textbf{smoothleft}: A fluid, smoothed sliding motion of the incoming shot from right to left with easing dynamics.
    \item \textbf{smoothright}: A fluid, smoothed sliding motion from left to right.
    \item \textbf{smoothup}: A fluid, smoothed vertical sliding motion from bottom to top.
    \item \textbf{smoothdown}: A fluid, smoothed vertical sliding motion from top to bottom.
    \item \textbf{circleopen}: Expands a circular aperture from the frame center to transition shots.
    \item \textbf{circleclose}: Contracts a circular aperture to hide the outgoing shot, revealing the underlying incoming shot.
    \item \textbf{vertopen}: Opens a vertical slit outward from the center horizontally to reveal the incoming shot.
    \item \textbf{vertclose}: Closes a vertical slit inward horizontally to transition between shots.
    \item \textbf{horzopen}: Opens a horizontal slit outward vertically from the center.
    \item \textbf{horzclose}: Closes a horizontal slit inward vertically.
    \item \textbf{dissolve}: A specialized pixel-level cross-dissolve that smooths the structural blending of overlapping frames.
    \item \textbf{pixelize}: Applies a highly blocky, pixelation filter that interpolates spatial frequencies between the two shots.
    \item \textbf{diagtl}: A diagonal wipe originating strictly from the top-left corner.
    \item \textbf{diagtr}: A diagonal wipe originating strictly from the top-right corner.
    \item \textbf{diagbl}: A diagonal wipe originating strictly from the bottom-left corner.
    \item \textbf{diagbr}: A diagonal wipe originating strictly from the bottom-right corner.
    \item \textbf{hlslice}: Interleaves horizontal slices sliding inward from the left boundary.
    \item \textbf{hrslice}: Interleaves horizontal slices sliding inward from the right boundary.
    \item \textbf{vuslice}: Interleaves vertical slices sliding inward from the top boundary.
    \item \textbf{vdslice}: Interleaves vertical slices sliding inward from the bottom boundary.
    \item \textbf{hblur}: Applies a heavy horizontal motion blur simultaneously during the temporal crossfade.
    \item \textbf{fadegrays}: Desaturates the outgoing shot to grayscale before temporally fading into the incoming color shot.
    \item \textbf{wipetl}: A corner-based linear wipe progressing towards the top-left.
    \item \textbf{wipetr}: A corner-based linear wipe progressing towards the top-right.
    \item \textbf{wipebl}: A corner-based linear wipe progressing towards the bottom-left.
    \item \textbf{wipebr}: A corner-based linear wipe progressing towards the bottom-right.
    \item \textbf{squeezeh}: Horizontally compresses and squeezes the outgoing shot to reveal the incoming one.
    \item \textbf{squeezev}: Vertically compresses and squeezes the outgoing shot.
    \item \textbf{zoomin}: Scales and zooms into the center of the outgoing shot while fading into the incoming shot.
    \item \textbf{fadefast}: A non-linear, temporally accelerated fade transition.
    \item \textbf{fadeslow}: A non-linear, temporally decelerated (eased) fade transition.
    \item \textbf{hlwind}: A horizontal wind-like pixel dispersion effect sweeping from the left.
    \item \textbf{hrwind}: A horizontal wind-like pixel dispersion effect sweeping from the right.
    \item \textbf{vuwind}: A vertical wind-like pixel dispersion effect sweeping upwards.
    \item \textbf{vdwind}: A vertical wind-like pixel dispersion effect sweeping downwards.
    \item \textbf{coverleft}: The incoming shot completely covers the outgoing shot by moving leftwards.
    \item \textbf{coverright}: The incoming shot covers the outgoing shot by moving rightwards.
    \item \textbf{coverup}: The incoming shot covers the outgoing shot by moving upwards.
    \item \textbf{coverdown}: The incoming shot covers the outgoing shot by moving downwards.
    \item \textbf{revealleft}: The outgoing shot continuously slides left to reveal the underlying incoming shot.
    \item \textbf{revealright}: The outgoing shot slides right to reveal the incoming shot.
    \item \textbf{revealup}: The outgoing shot slides up to reveal the incoming shot.
    \item \textbf{revealdown}: The outgoing shot slides down to reveal the incoming shot.
\end{enumerate}

\subsection{Extended Details of Training and Inference}
\label{sec:supp_training}

\subsubsection{Training Dataset Statistics and Processing}
As detailed in Table 1 of the main text, our finalized training dataset comprises over 233,000 videos totaling approximately 1,563 hours. It features more than 690,000 explicitly annotated transitions across diverse temporal dynamics (e.g., abrupt cuts, normal gradual transitions, and prolonged transitions). To maintain stable memory usage and a consistent temporal context during the VLM optimization process, all video inputs sourced from both the data engine and the public datasets are uniformly chunked into clips of $\leq 10$ seconds and sampled at a fixed rate of 25 FPS.

\begin{figure}[htbp]
    \centering
    \includegraphics[width=0.9\textwidth]{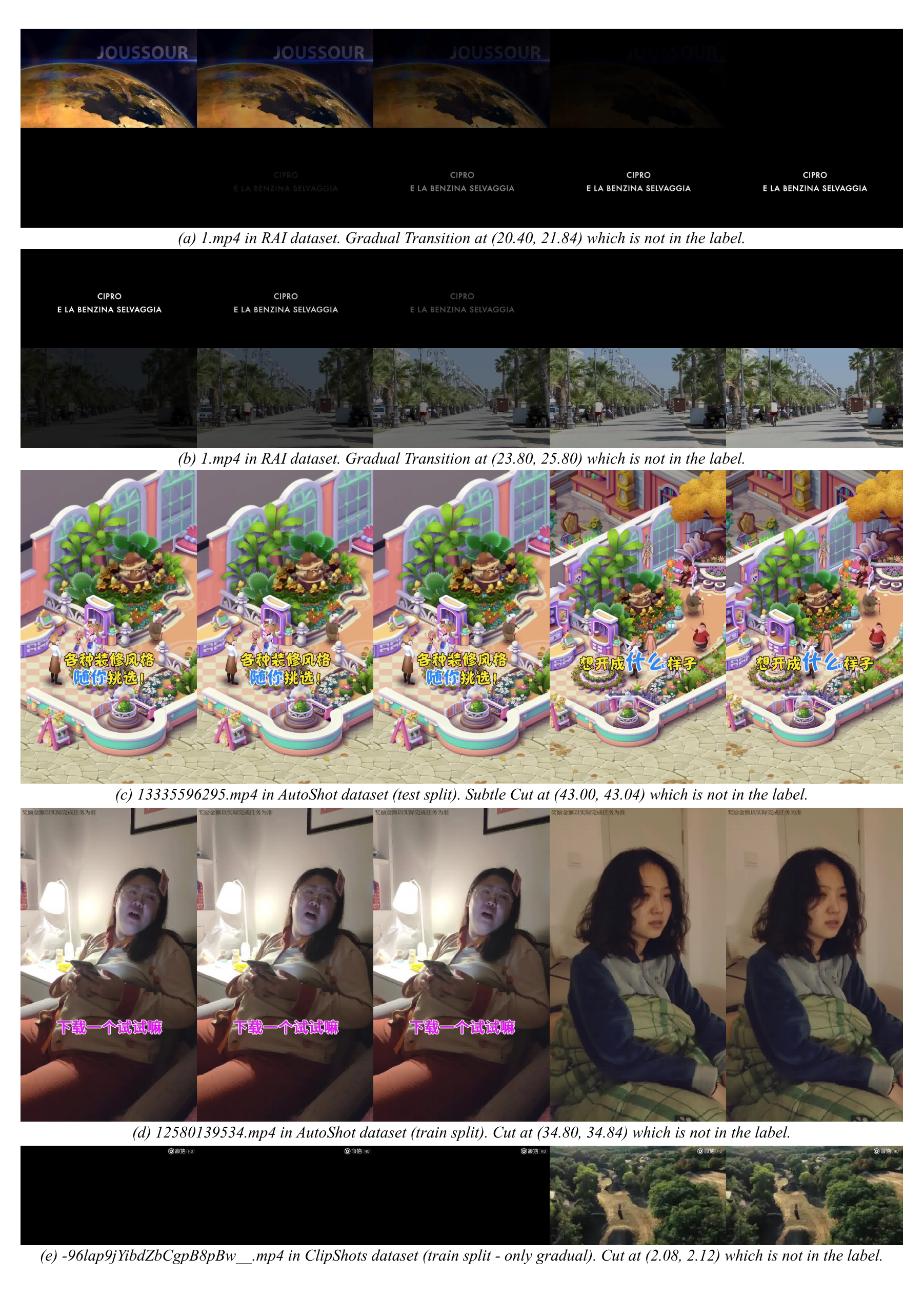}
    \caption{\textbf{Visualizations of omitted transitions in legacy public datasets (Part A).} We uniformly sample 5 or 10 frames around the exact temporal locations of the transitions. Despite the clear visual evidence of shot changes, these transitions are entirely absent from the original ground-truth labels.}
    \label{fig:appendix_bad_cases_a}
\end{figure}

\begin{figure}[htbp]
    \centering
    \includegraphics[width=0.9\textwidth]{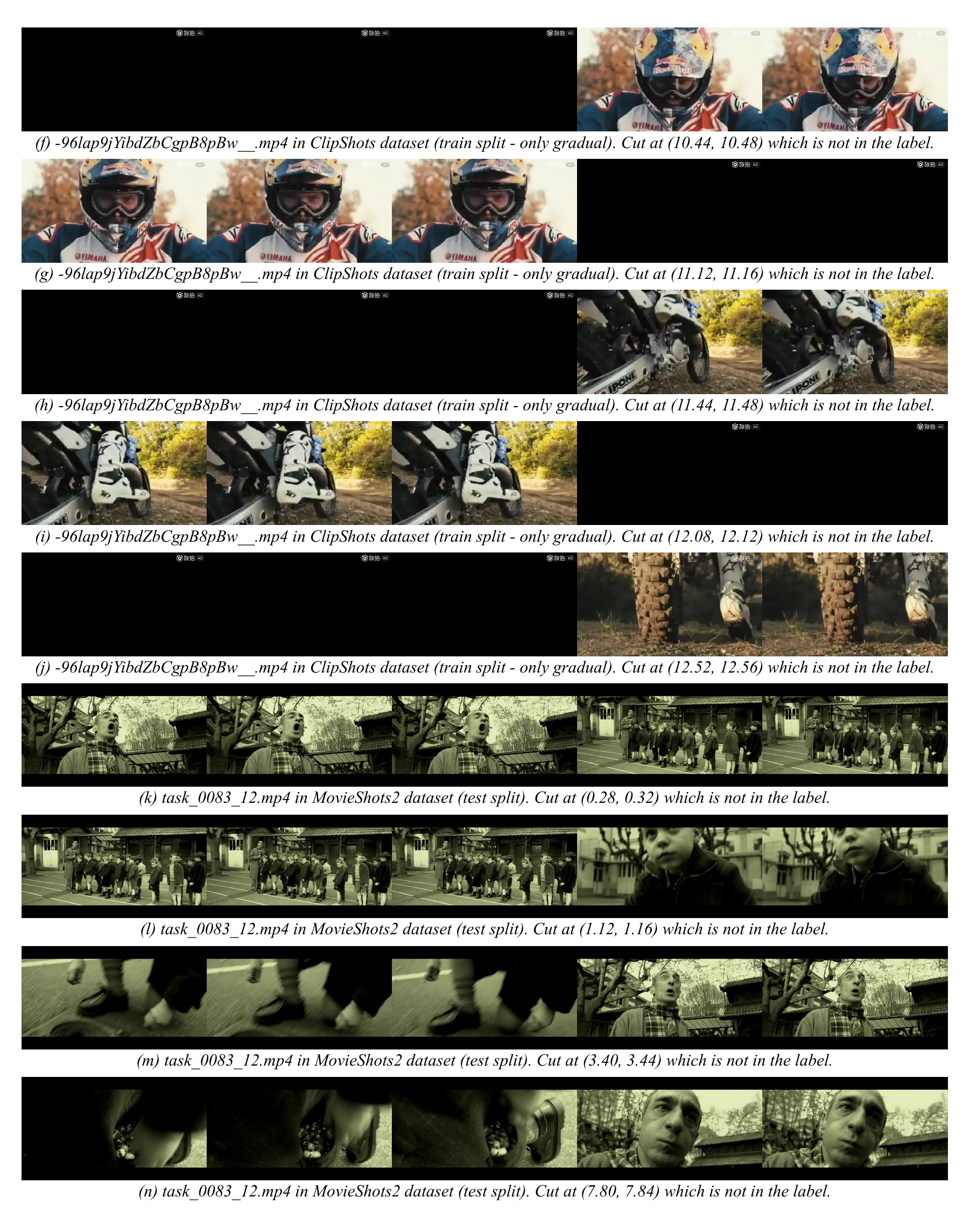}
    \caption{\textbf{Visualizations of omitted transitions in legacy public datasets (Part B).} We uniformly sample 5 or 10 frames around the exact temporal locations of the transitions. Despite the clear visual evidence of shot changes, these transitions are entirely absent from the original ground-truth labels.}
    \label{fig:appendix_bad_cases_b}
\end{figure}

\subsubsection{Quality-Aware Sampling Tiers}
To implement the quality-aware mixed sampling strategy, we categorized all constituent datasets into discrete quality tiers (e.g., Very High, High, Medium) based on extensive manual sampling evaluations. This hierarchical categorization ensures that datasets with highly accurate, segment-level annotations have a proportionally higher probability of being sampled in each training batch compared to those containing noisier legacy annotations.

\subsubsection{Sliding-Window Inference Parameters}
During the arbitrary video inference phase, a given video is continuously partitioned into overlapping temporal windows. We mathematically define the window size as $W$ and the temporal stride as $S$, where $S \leq W$. This explicitly enforces a temporal overlap of $W - S$ between adjacent windows (in practice, $W=10$s and $S=9$s, leaving a 1-second overlap). For each local window, the model generates textual outputs representing the start and end timestamps of any detected transitions relative to that specific window's timeline. These local timestamps are subsequently projected back onto the global video timeline before we apply temporal Non-Maximum Suppression (NMS) to merge overlapping segments.

\subsubsection{Detailed Training Configurations}
To effectively optimize \ccmethod{} for the continuous Shot Transition Detection (STD) task, we implemented our training pipeline based on the open-source VeOmni framework. The training details and specific hyperparameter configurations are comprehensively described below:

\noindent\textbf{Optimization and Learning Rate.} The model is initialized with the Qwen3-VL-4B-Instruct weights. Critically, we leave the Vision Transformer (ViT) entirely unfrozen (\texttt{freeze\_vit: false}) to enable full-parameter fine-tuning, allowing the vision encoder to smoothly adapt to the newly fused color and optical flow modalities utilizing our zero-padding initialization strategy. We optimize the network using the AdamW optimizer with a maximum gradient norm clipped at $1.0$. The peak learning rate is strictly set to $1.0 \times 10^{-5}$, modulated by a cosine learning rate decay schedule to ensure stable convergence. 

\noindent\textbf{Batching and Sequence Length.} The maximum sequence length during training is deliberately set to $16,384$ tokens. This extended context window comfortably accommodates the dense visual tokens extracted from the maximum $10.0$-second training clips, alongside the customized text prompts. The training is conducted across 8 NVIDIA H100 (80GB) GPUs. We assign a micro-batch size of $4$ per device, yielding an effective global batch size of $32$. The model is fine-tuned for a maximum of $500$ optimization steps.

\noindent\textbf{Memory and Computational Efficiency.} Given the intensive memory demands of processing high-frame-rate multimodal inputs, we integrate several advanced memory-optimization mechanisms. Specifically, we utilize Fully Sharded Data Parallel 2 (FSDP2) for distributed training, coupled with standard gradient checkpointing. To accelerate the attention computation, \texttt{flash\_attention\_2} is explicitly adopted. Furthermore, we enable padding-free training relying on position IDs (\texttt{rmpad\_with\_pos\_ids: true}) combined with a dynamic batch size buffer, which drastically minimizes redundant computations on zero-padded tokens.

\noindent\textbf{Prompt Formatting.} To align the VLM's generative output with the strict segment-level evaluation requirements of the STD task, our customized user prompt explicitly instructs the model to output the detected transition information exclusively as a single, raw JSON array of temporal tuples. The generation of markdown wrappers (e.g., \texttt{```json}) or any auxiliary explanatory text is strictly prohibited during the supervised fine-tuning phase.

\subsection{Bad Annotations in Public Datasets}
\label{sec:appendix_bad_annotations}

During our data inspection and re-annotation process, we identified a general issue regarding the annotation fidelity of public datasets designed for SBD task, spanning both training and testing splits. This detrimental issue is the omission of valid transition boundaries---i.e., some shot transitions are completely unrecorded in the original ground-truth labels. This inherently high false-negative rate in the annotations significantly compromises the reliability of any evaluation conducted solely on these datasets, as models correctly identifying these unannotated transitions would be unfairly penalized. \textbf{However, given the large scale of the training data, exhaustively re-annotating every single video is practically infeasible. Consequently, the inevitable presence of these residual noisy annotations in the training splits may introduce adverse gradients and negatively impact the model optimization process.} This practical limitation further underscores the necessity of the quality-aware mixed sampling strategy we implemented during training.

To demonstrate this critical flaw, Figure \ref{fig:appendix_bad_cases_a} and \ref{fig:appendix_bad_cases_b} presents a representative selection of omitted transitions randomly discovered during our manual review of the public datasets. For optimal visual clarity, we uniformly sample 5 or 10 continuous frames within the immediate temporal neighborhood of each missed transition. As unequivocally shown in the visualization, these sequences contain distinct and visually obvious shot changes (both abrupt cuts and gradual transitions) that were entirely overlooked by the original human annotators. This further validates the absolute necessity of our meticulously re-annotated TBD Benchmark.

\subsection{More Details of Quantitative Experiments}

\begin{figure}[h]
	\centering
	\includegraphics[width=0.43\linewidth]{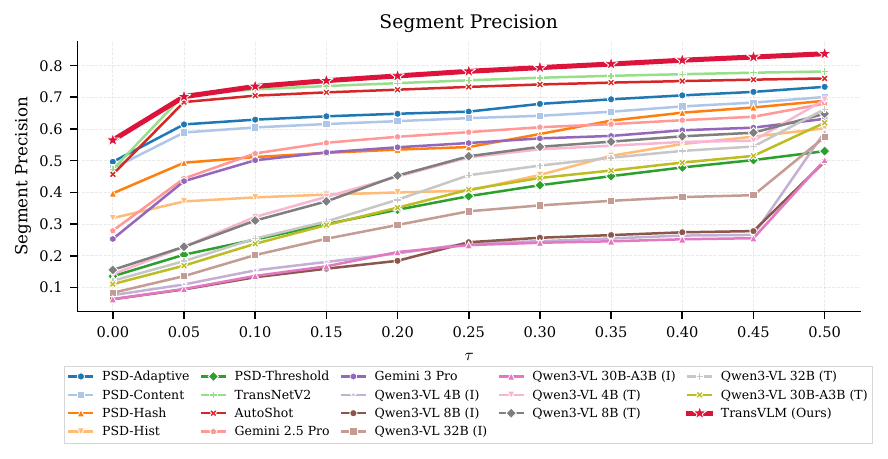}
	\hfill
	\includegraphics[width=0.43\linewidth]{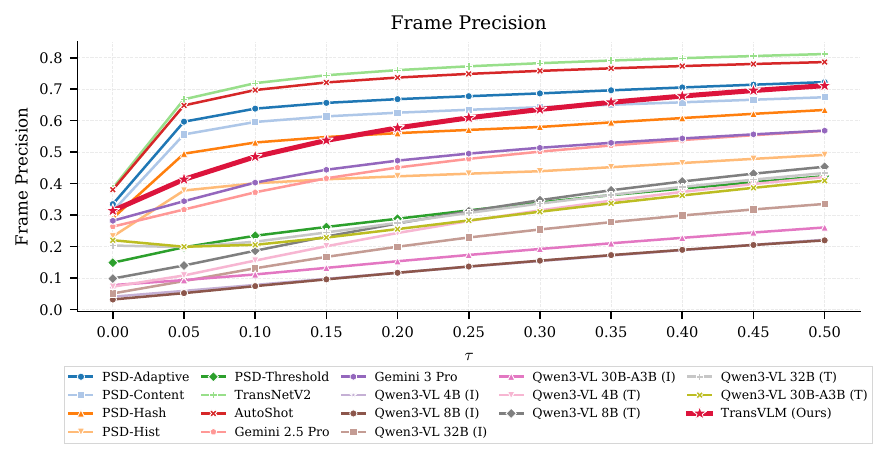}
	    
    \vspace{2em}
    
    \includegraphics[width=0.43\linewidth]{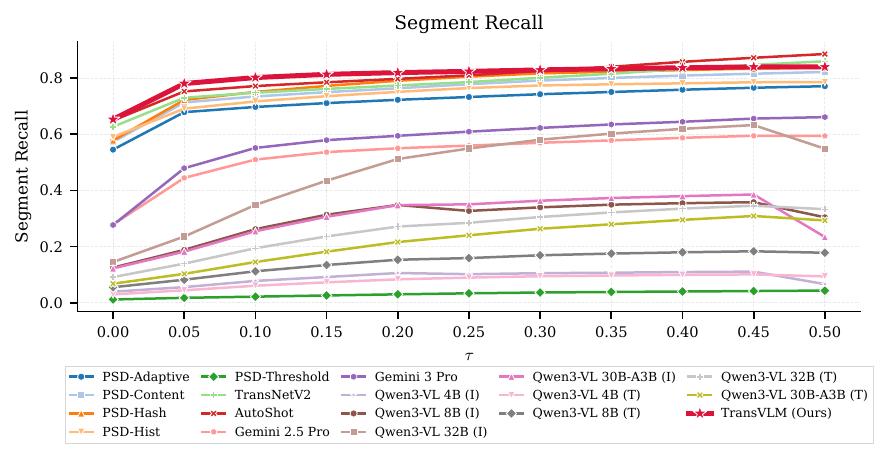}
	\hfill
	\includegraphics[width=0.43\linewidth]{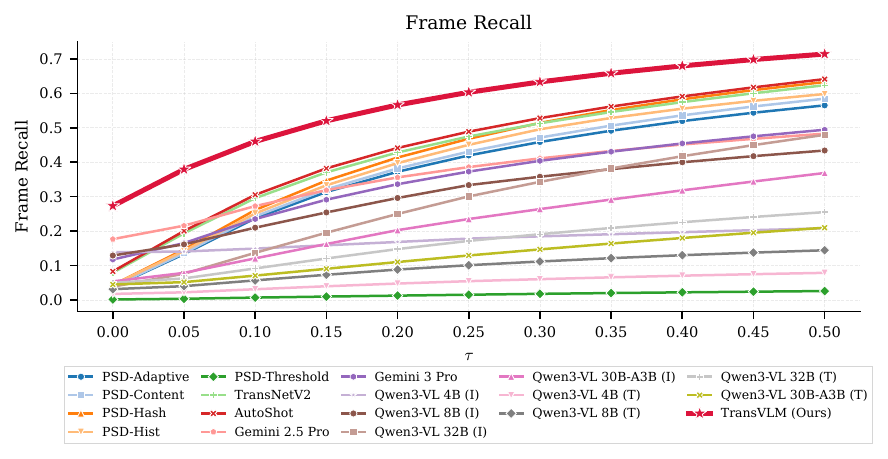}
		    
    \vspace{2em}
    
    \includegraphics[width=0.43\linewidth]{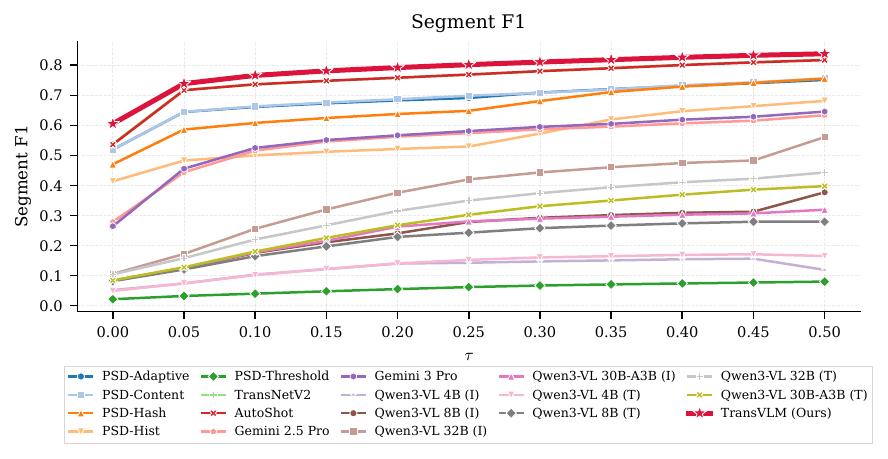}
	\hfill
	\includegraphics[width=0.43\linewidth]{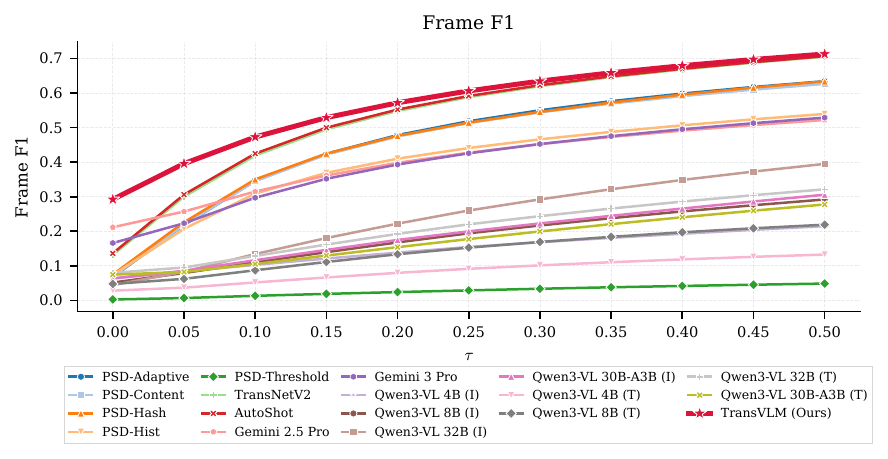}
		    
    \vspace{2em}
    
    \includegraphics[width=0.43\linewidth]{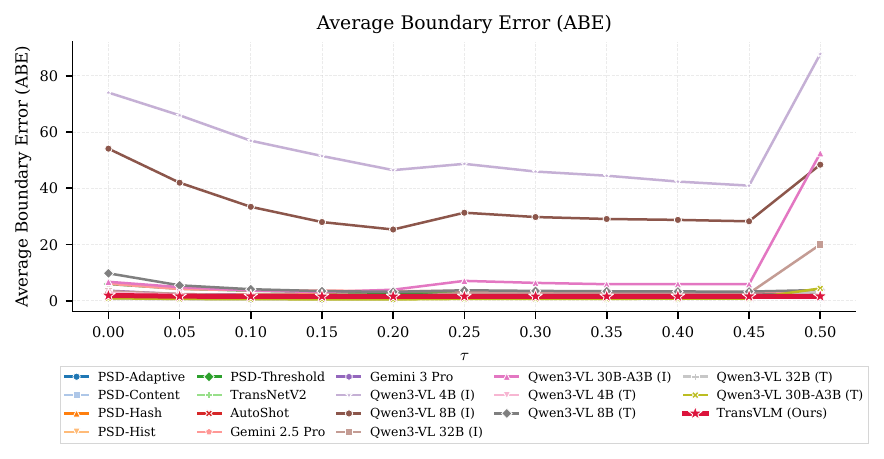}
    \caption{
    	\textbf{\textbf{Quantitative comparison visualization on all public data.}}
    }
    \label{fig:supp_expr_comp_pblc}
\end{figure}

\begin{figure}[t]
	\centering
	\includegraphics[width=0.48\linewidth]{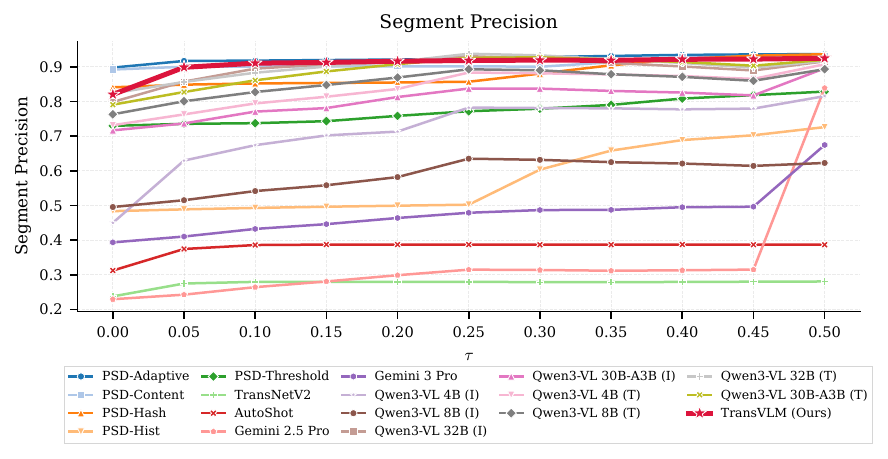}
	\hfill
	\includegraphics[width=0.48\linewidth]{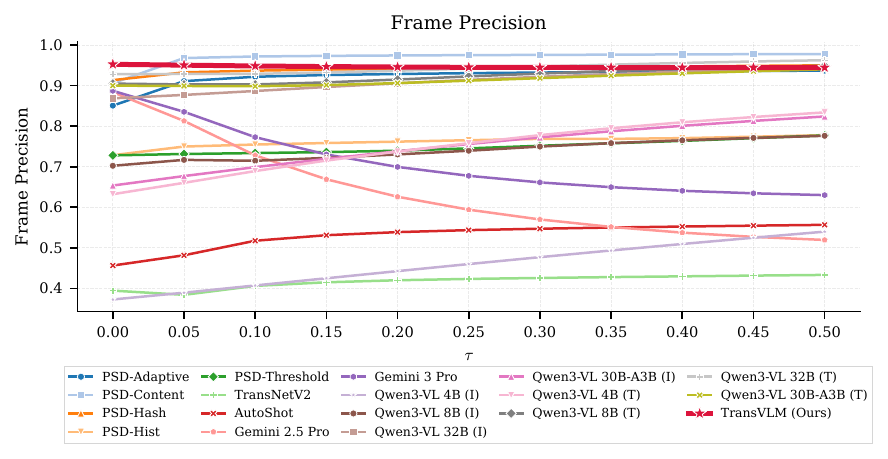}
	    
    \vspace{2em}
    
    \includegraphics[width=0.48\linewidth]{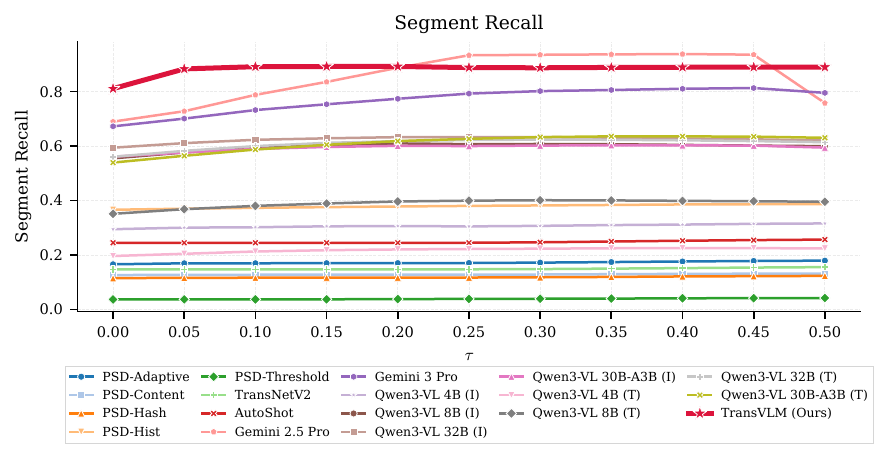}
	\hfill
	\includegraphics[width=0.48\linewidth]{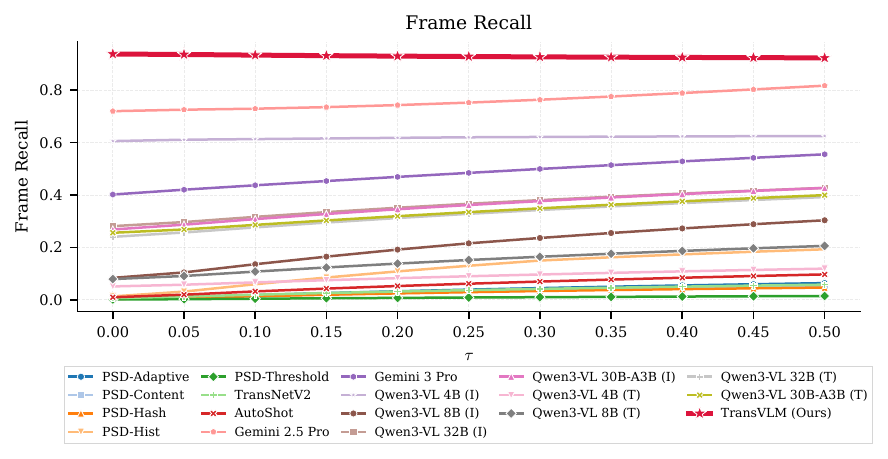}
		    
    \vspace{2em}
    
    \includegraphics[width=0.48\linewidth]{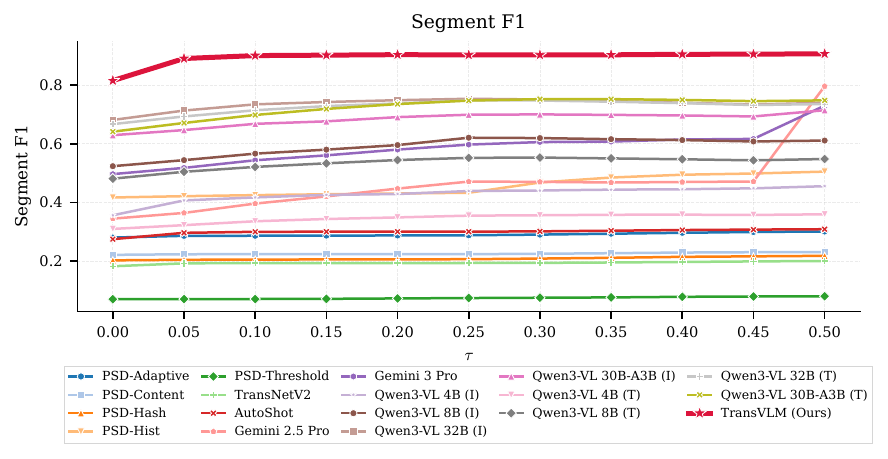}
	\hfill
	\includegraphics[width=0.48\linewidth]{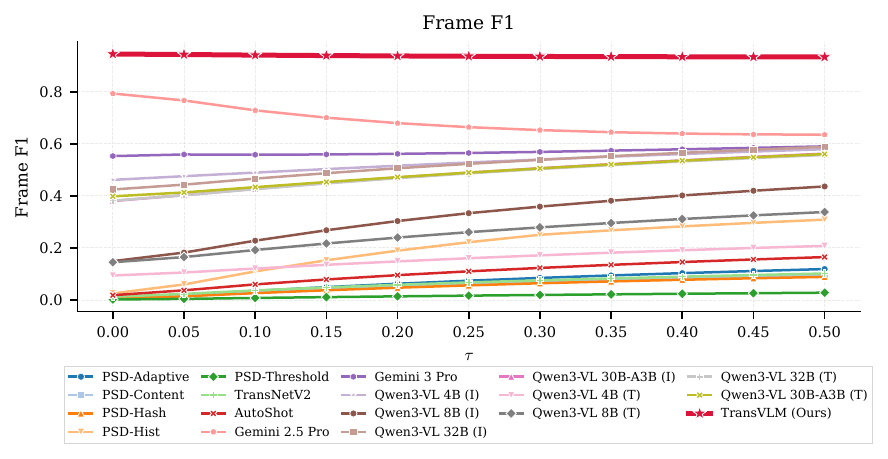}
		    
    \vspace{2em}
    
    \includegraphics[width=0.48\linewidth]{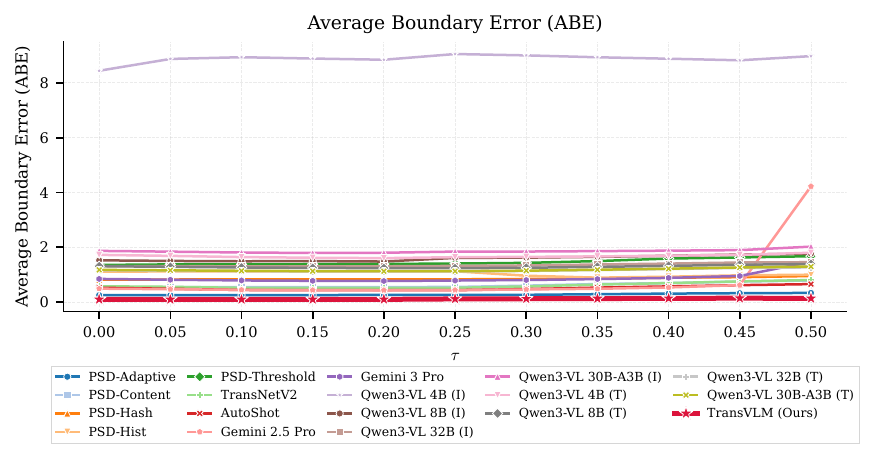}
    \caption{
    	\textbf{\textbf{Quantitative comparison visualization on all synthetic data.}}
    }
    \label{fig:supp_expr_comp_synthtc}
\end{figure}

\begin{figure}[t]
	\centering
	\includegraphics[width=0.48\linewidth]{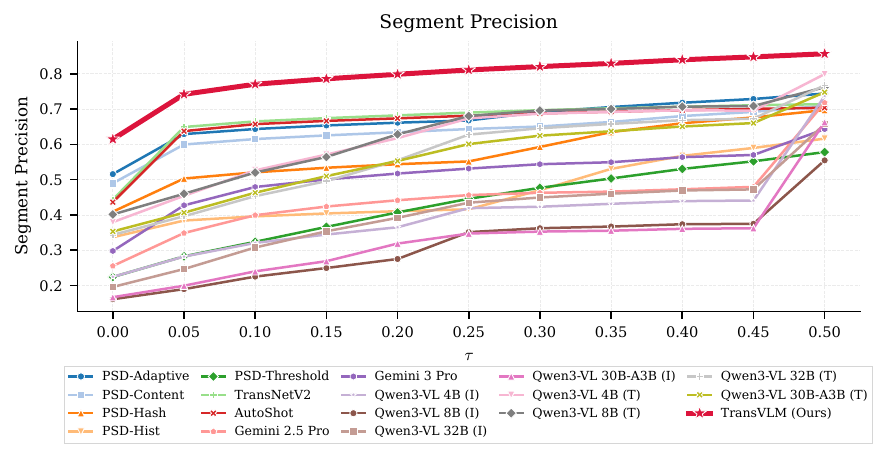}
	\hfill
	\includegraphics[width=0.48\linewidth]{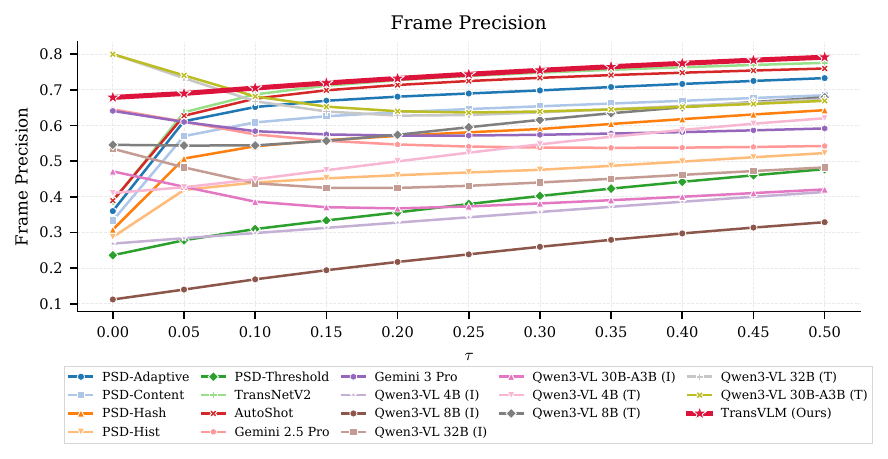}
	    
    \vspace{2em}
    
    \includegraphics[width=0.48\linewidth]{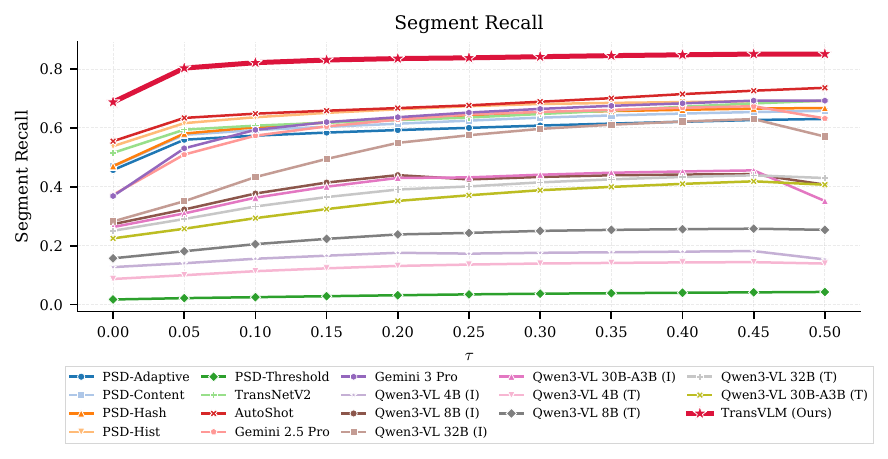}
	\hfill
	\includegraphics[width=0.48\linewidth]{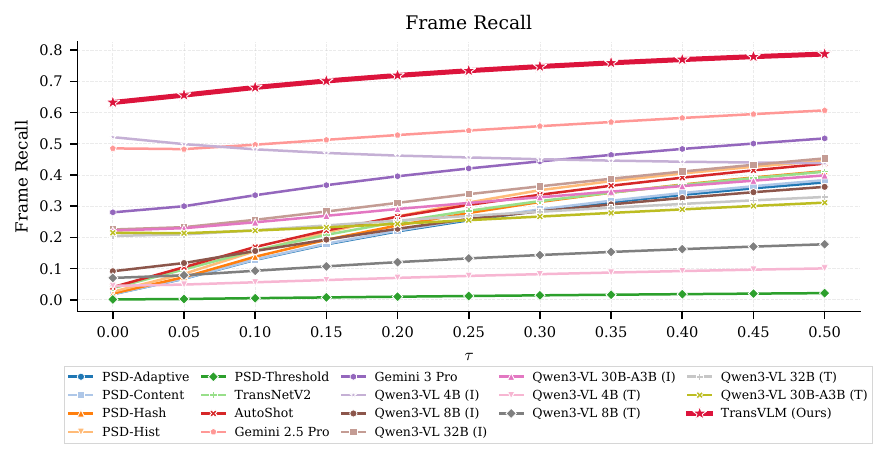}
		    
    \vspace{2em}
    
    \includegraphics[width=0.48\linewidth]{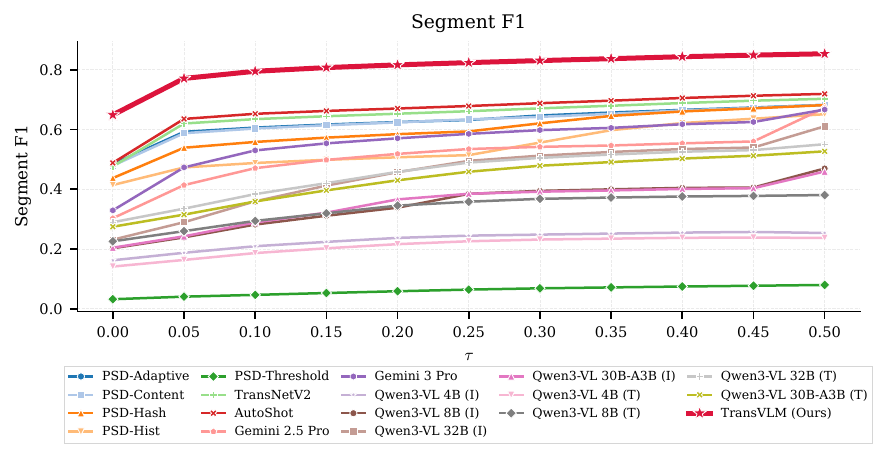}
	\hfill
	\includegraphics[width=0.48\linewidth]{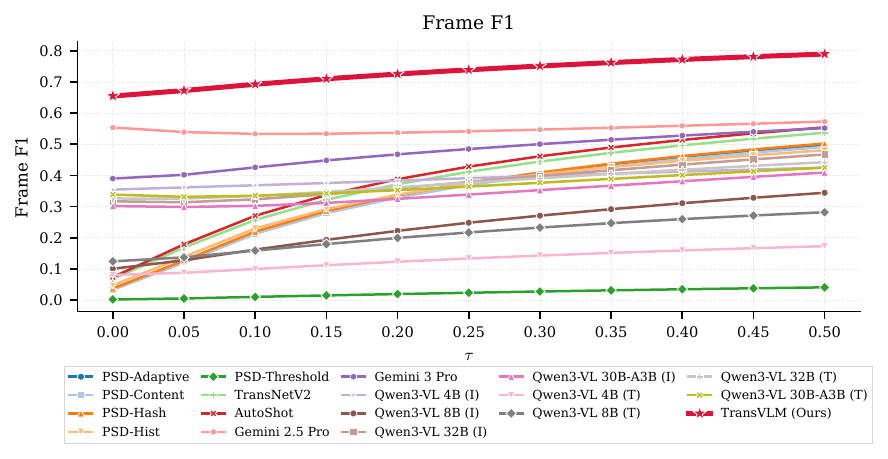}
		    
    \vspace{2em}
    
    \includegraphics[width=0.48\linewidth]{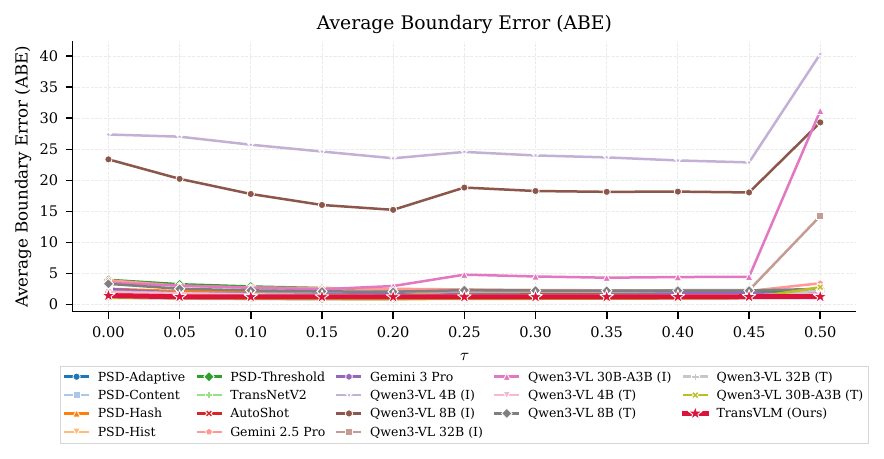}
    \caption{
    	\textbf{\textbf{Quantitative comparison visualization on all data.}}
    }
    \label{fig:supp_expr_comp_all}
\end{figure}

In this section, we provide an exhaustive breakdown of our quantitative evaluations. The following figures illustrate the performance curves of various methods as a function of the temporal tolerance $\tau$. These evaluations are systematically conducted across different aggregated data domains (public data \ref{fig:supp_expr_comp_pblc}, synthetic data \ref{fig:supp_expr_comp_synthtc}, and the overall benchmark \ref{fig:supp_expr_comp_all}). As explicitly demonstrated by the plotted curves, our proposed \ccmethod{} consistently achieves superior overall performance. Regardless of the strictness of the temporal tolerance, our framework robustly dominates the primary evaluation metrics---specifically the segment-level $F_1$ and frame-level $F_1$ scores---further validating its absolute effectiveness and stability on the continuous transition detection task.

The horizontal axis represents $\tau$, while the vertical axis represents the value of selected metric. 

%
%
%
%
%
%
%
%
%
%
%
%
%
%
%
%
%
%
%
%
%
%
%

\clearpage

\subsection{Extended Details of Ablation Studies}

\begin{figure}[h]
	\centering
	\includegraphics[width=0.43\linewidth]{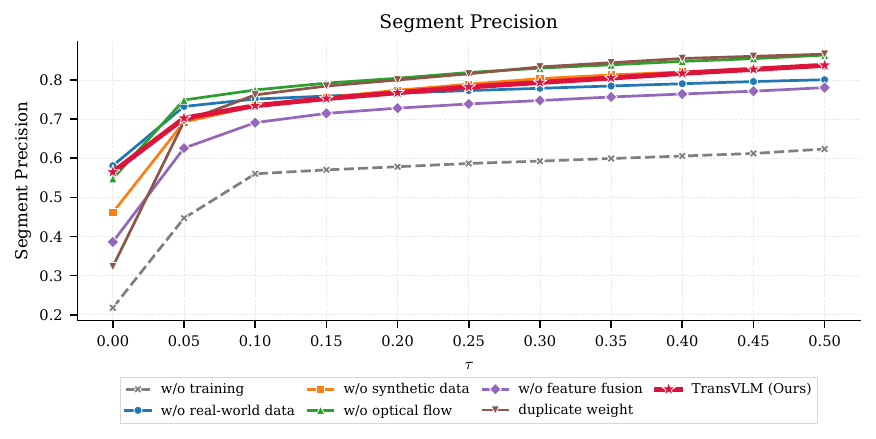}
	\hfill
	\includegraphics[width=0.43\linewidth]{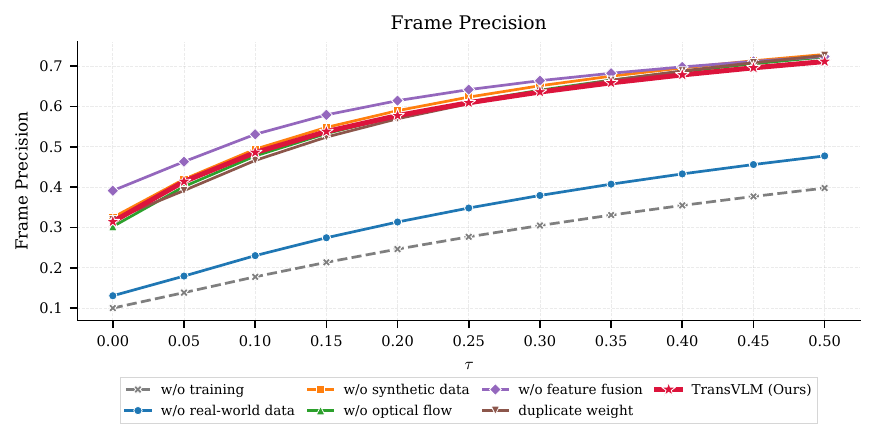}
	    
    \vspace{2em}
    
    \includegraphics[width=0.43\linewidth]{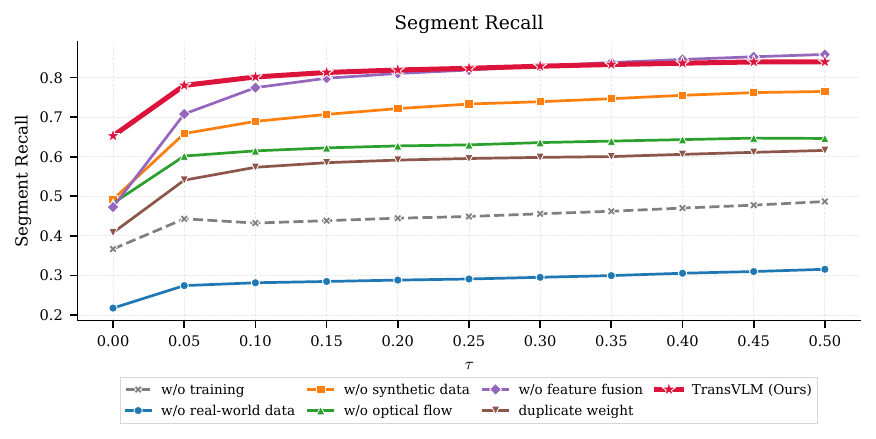}
	\hfill
	\includegraphics[width=0.43\linewidth]{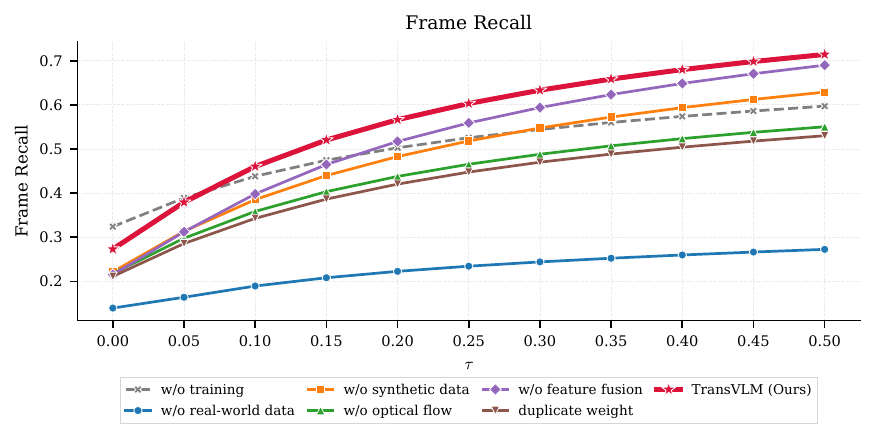}
		    
    \vspace{2em}
    
    \includegraphics[width=0.43\linewidth]{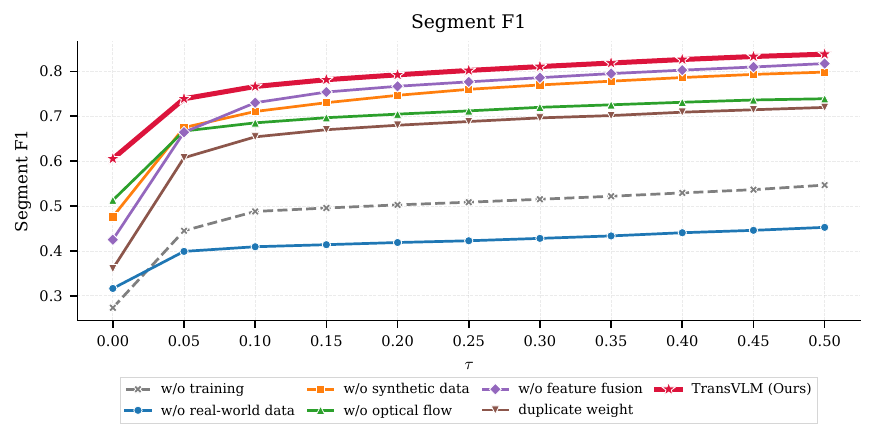}
	\hfill
	\includegraphics[width=0.43\linewidth]{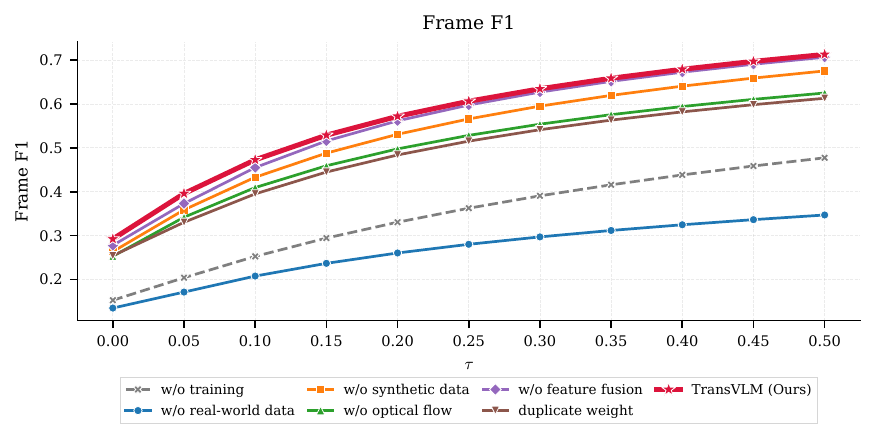}
		    
    \vspace{2em}
    
    \includegraphics[width=0.43\linewidth]{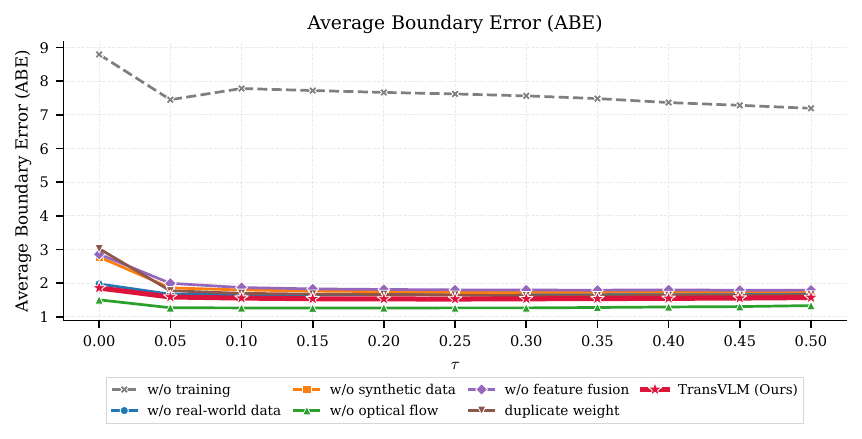}
    \caption{
    	\textbf{\textbf{Ablation study visualization on all public data.}}
    }
    \label{fig:supp_expr_abla_pblc}
\end{figure}

\begin{figure}[t]
	\centering
	\includegraphics[width=0.48\linewidth]{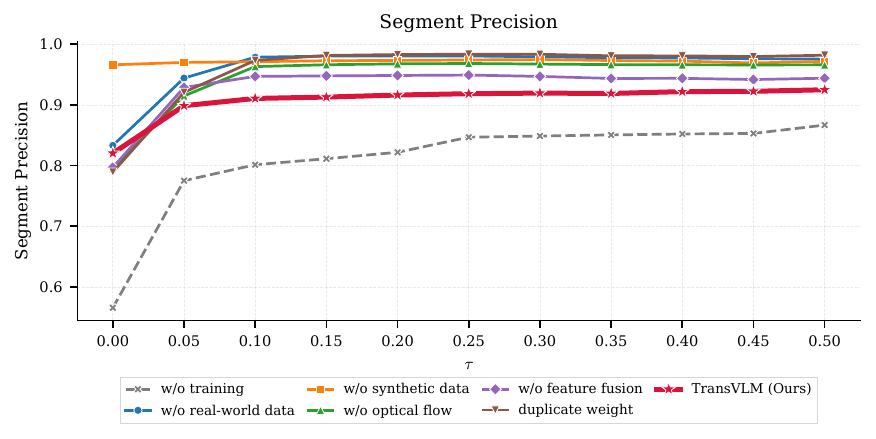}
	\hfill
	\includegraphics[width=0.48\linewidth]{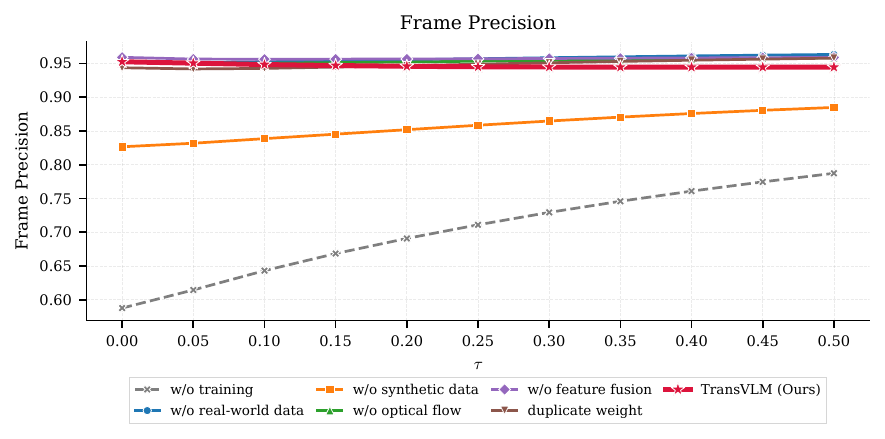}
	    
    \vspace{2em}
    
    \includegraphics[width=0.48\linewidth]{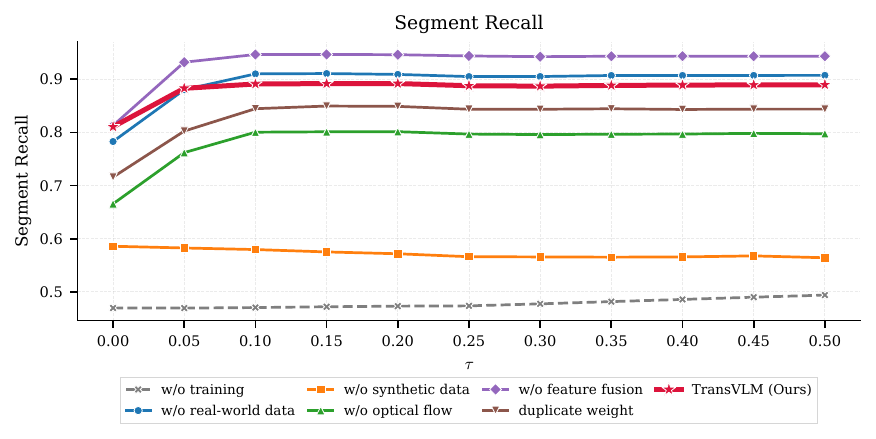}
	\hfill
	\includegraphics[width=0.48\linewidth]{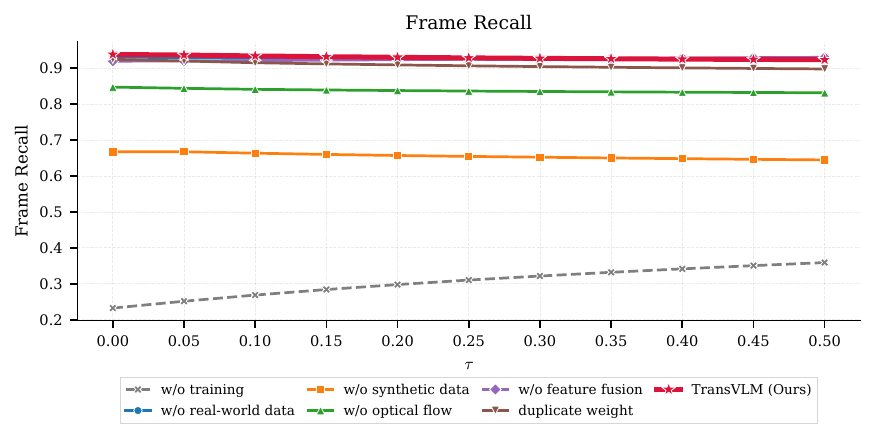}
		    
    \vspace{2em}
    
    \includegraphics[width=0.48\linewidth]{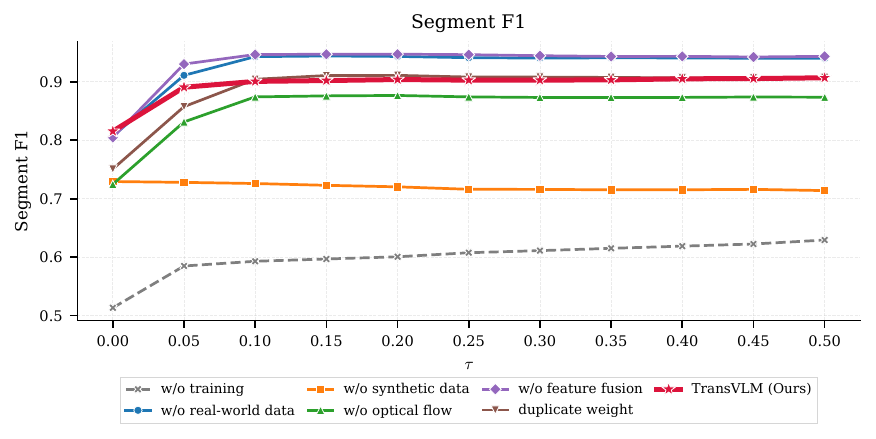}
	\hfill
	\includegraphics[width=0.48\linewidth]{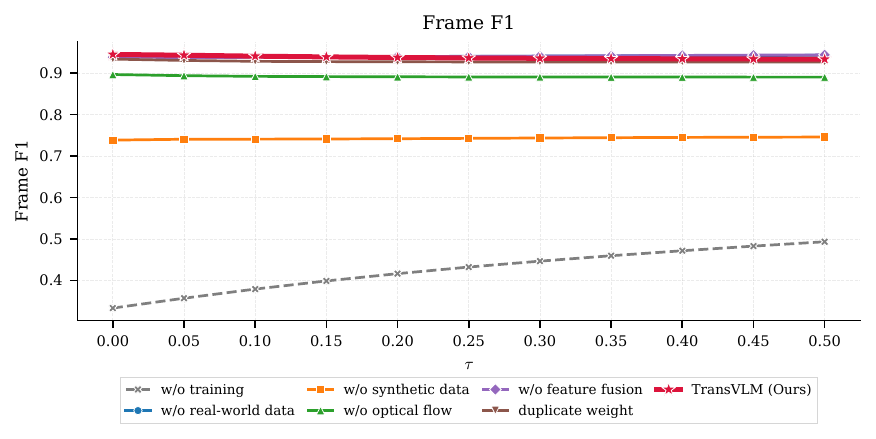}
		    
    \vspace{2em}
    
    \includegraphics[width=0.48\linewidth]{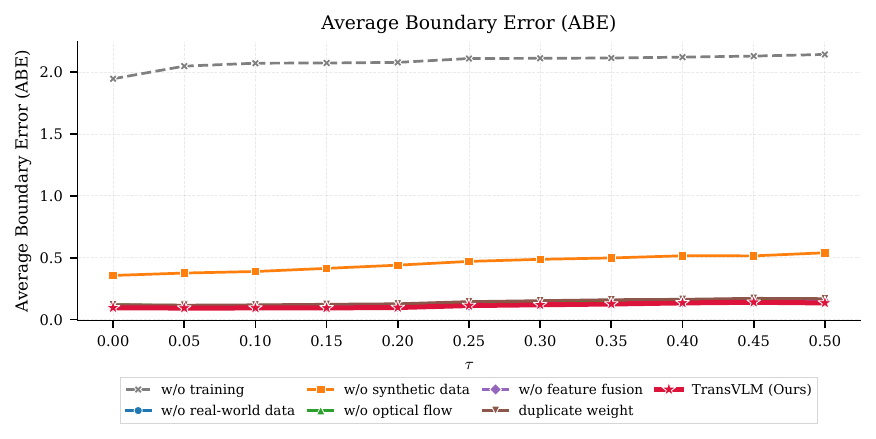}
    \caption{
    	\textbf{\textbf{Ablation study visualization on all synthetic data.}}
    }
    \label{fig:supp_expr_abla_synthtc}
\end{figure}

\begin{figure}[t]
	\centering
	\includegraphics[width=0.48\linewidth]{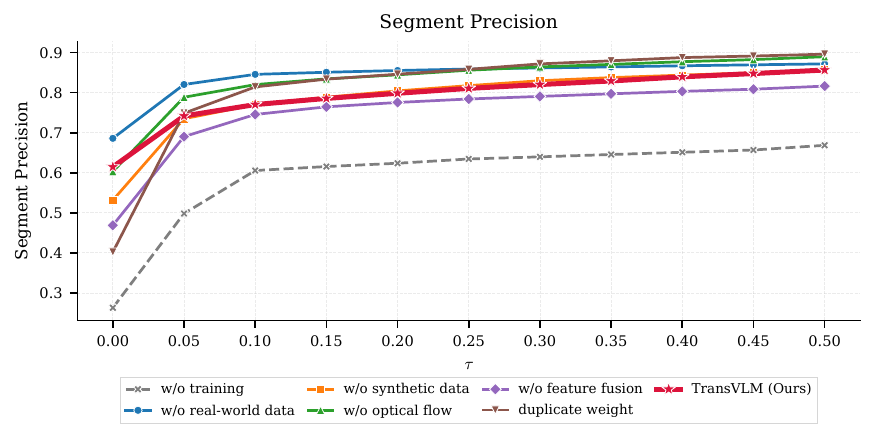}
	\hfill
	\includegraphics[width=0.48\linewidth]{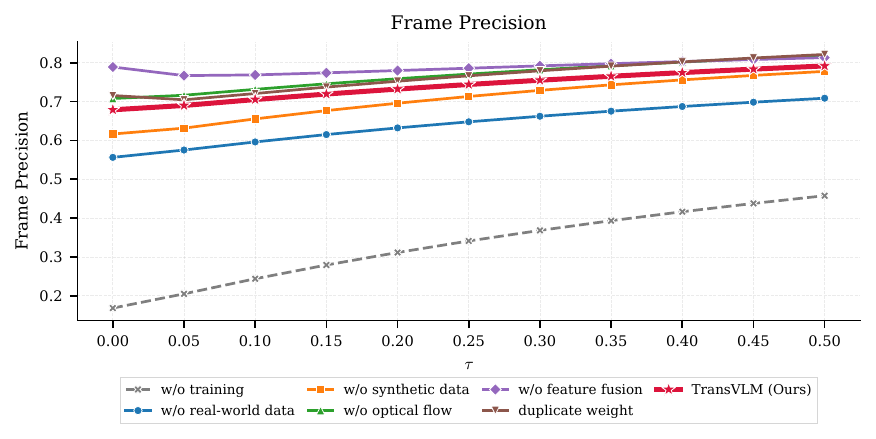}
	    
    \vspace{2em}
    
    \includegraphics[width=0.48\linewidth]{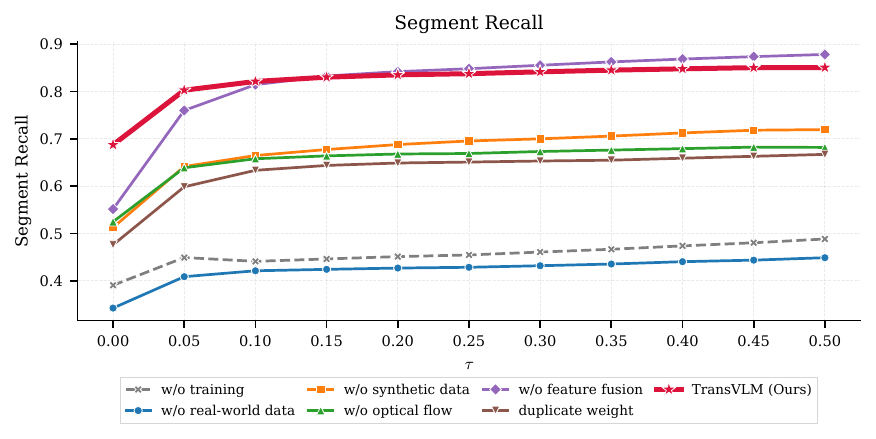}
	\hfill
	\includegraphics[width=0.48\linewidth]{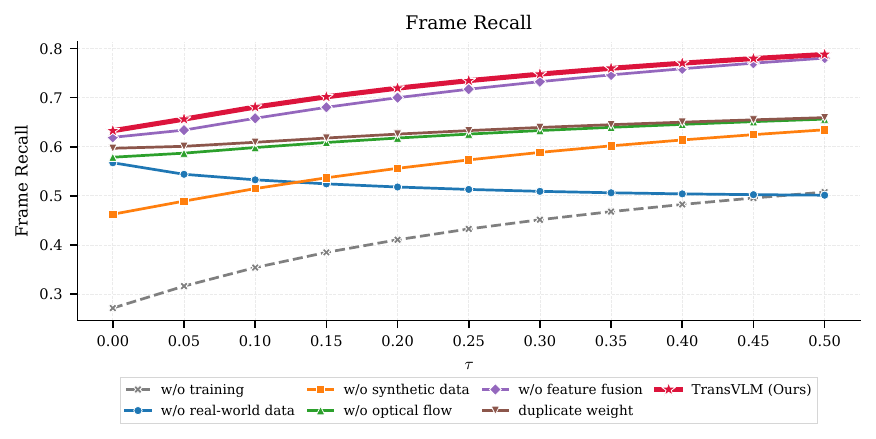}
		    
    \vspace{2em}
    
    \includegraphics[width=0.48\linewidth]{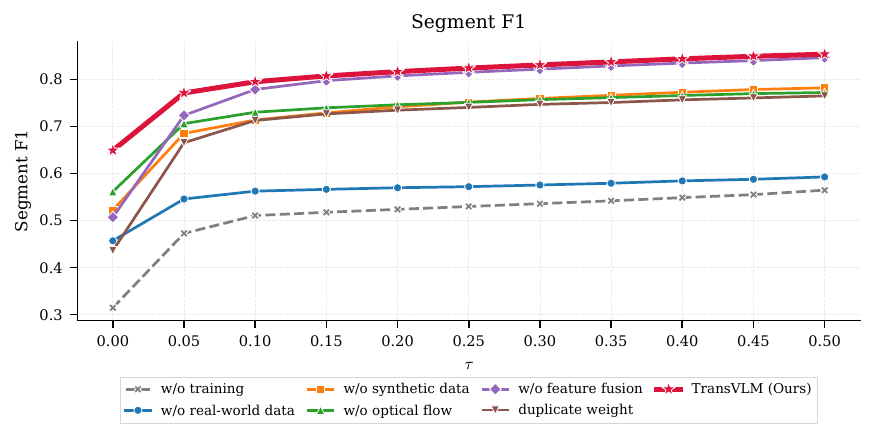}
	\hfill
	\includegraphics[width=0.48\linewidth]{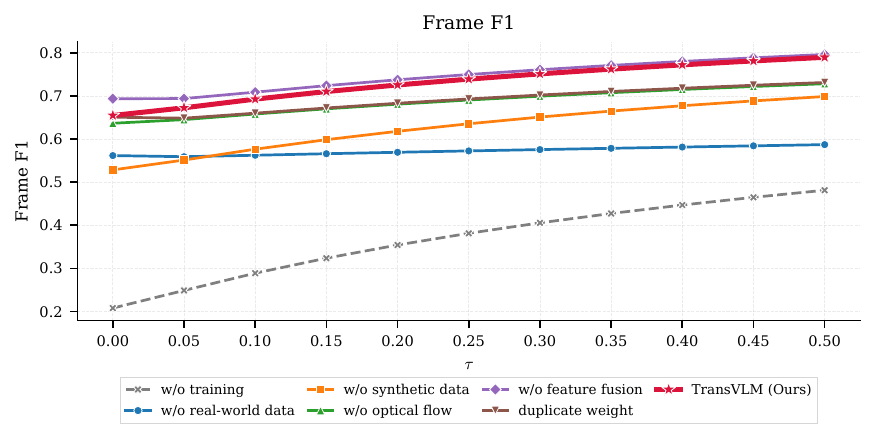}
		    
    \vspace{2em}
    
    \includegraphics[width=0.48\linewidth]{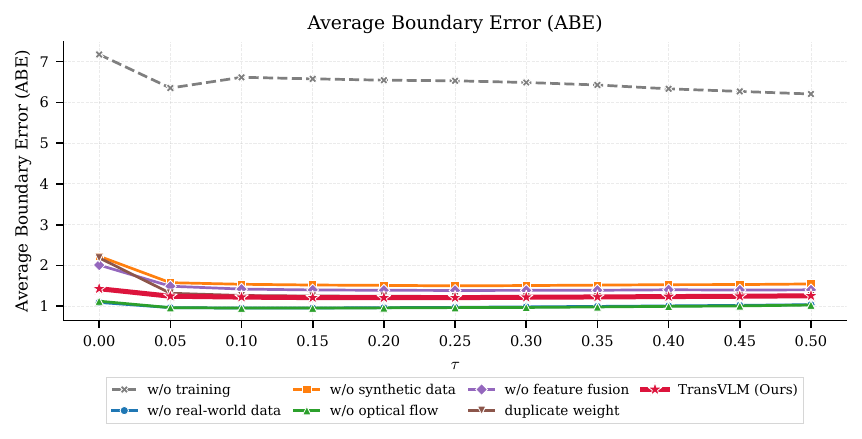}
    \caption{
    	\textbf{\textbf{Ablation study visualization on all data.}}
    }
    \label{fig:supp_expr_abla_all}
\end{figure}

In this section, we provide an exhaustive breakdown of our comprehensive ablation studies to further validate the individual contributions of each component within \ccmethod{}. The following figures illustrate the performance curves of various methods as a function of the temporal tolerance $\tau$. These evaluations are systematically conducted across different aggregated data domains (public data \ref{fig:supp_expr_abla_pblc}, synthetic data \ref{fig:supp_expr_abla_synthtc}, and the overall benchmark \ref{fig:supp_expr_abla_all}). As explicitly demonstrated by the plotted curves, our proposed \ccmethod{} consistently achieves superior balance between public and synthetic data.

The horizontal axis represents $\tau$, while the vertical axis represents the value of selected metric. 

%
%
%
%
%
%
%
%
%
%
%
%
%
%
%
%
%
%
%
%
%
%
%

\end{document}